%% file: main.tex
\pdfoutput=1

\documentclass[11pt]{article}

\usepackage{acl}
\usepackage[subtle]{savetrees}

\usepackage{times}
\usepackage{latexsym}

\usepackage[T1]{fontenc}

\usepackage[utf8]{inputenc}

\usepackage{microtype}

\usepackage[english]{babel}
\usepackage{amsmath}
\usepackage{graphicx}
\usepackage{xcolor}
\usepackage{lipsum}
\usepackage{amssymb}
\usepackage{dsfont}
\usepackage{mathtools}
\usepackage{amsthm}
\usepackage{multirow}
\usepackage{tikz}
\usepackage{enumitem}
\usepackage{hwemoji}

\graphicspath{{figures}}

\usepackage{xspace}
\newcommand{\myparagraph}[1]{\smallskip\noindent\textbf{#1}\xspace}

\newtheorem{theorem}{Theorem}
\theoremstyle{definition}
\newtheorem{definition}{Definition}[section]

\DeclareRobustCommand{\berthatexplainsquare}{
  \tikz[baseline=-0.5ex]\draw[black,fill=hatexplain] (0,-0.07) -- (0.2,-0.07) -- (0.1,0.13) -- cycle;
}
\DeclareRobustCommand{\ourscircle}{
  \tikz[baseline=-0.5ex]\fill[ours] (0,0) circle (0.08);
}

\title{Exploring the Trade-off Between Model Performance and Explanation Plausibility of Text Classifiers Using Human Rationales}

\author{
    Lucas E. Resck\textnormal{\textsuperscript{1}} \and Marcos M. Raimundo\textnormal{\textsuperscript{2}} \and Jorge Poco\textnormal{\textsuperscript{1}} \\
    \textsuperscript{1}Fundação Getulio Vargas, Rio de Janeiro, Brazil \\
    \textsuperscript{2}Universidade Estadual de Campinas (UNICAMP), Campinas, Brazil \\
    \texttt{lucas.resck@fgv.br}, \texttt{mraimundo@ic.unicamp.br}, \texttt{jorge.poco@fgv.br}
}

\begin{document}
    \maketitle
    
    \begin{tikzpicture}[overlay, remember picture]
        \node[anchor=south, font=\footnotesize, yshift=0.6cm] at (current page.south) { 
            Copyright © 2024 Association for Computational Linguists (ACL). 
            Licensed under the \href{https://creativecommons.org/licenses/by/4.0/}{CC BY 4.0} license. LaTeX source file compiled independently.
        };    
        \node[anchor=south, font=\footnotesize, yshift=0.2cm] at (current page.south) { 
            \textbf{Disclaimer:} No warranties are given. The ACL shall not be liable for any damages arising from the use of this article.
        };
    \end{tikzpicture}

    \begin{abstract}
        \input{abstract}
    \end{abstract}

    \input{content}

    \bibliography{main}

    \appendix
    \input{appendix}
\end{document}

%% file: abstract.tex
Saliency post-hoc explainability methods are important tools for understanding increasingly complex NLP models. While these methods can reflect the model's reasoning, they may not align with human intuition, making the explanations not plausible. 
In this work, we present a methodology for incorporating rationales, which are text annotations explaining human decisions, into text classification models. This incorporation enhances the plausibility of post-hoc explanations while preserving their faithfulness. Our approach is agnostic to model architectures and explainability methods.
We introduce the rationales during model training by augmenting the standard cross-entropy loss with a novel loss function inspired by contrastive learning. By leveraging a multi-objective optimization algorithm, we explore the trade-off between the two loss functions and generate a Pareto-optimal frontier of models that balance performance and plausibility.
Through extensive experiments involving diverse models, datasets, and explainability methods, we demonstrate that our approach significantly enhances the quality of model explanations without causing substantial (sometimes negligible) degradation in the original model's performance.\footnote{Code and data are available at \url{https://github.com/visual-ds/plausible-nlp-explanations}.}

%% file: content.tex
\input{content/introduction.tex}

\input{content/related_work.tex}

\input{content/theoretical_background.tex}

\input{content/methodology.tex}

\input{content/experiments.tex}

\input{content/discussion.tex}

\input{content/conclusion.tex}

\input{content/limitations.tex}

\input{content/ethics_statement.tex}

\input{content/acknowledgements.tex}

%% file: content/introduction.tex
\section{Introduction}
    \label{sec:introduction}

    The complexity of text classification models and architectures has recently grown, posing challenges in comprehending the rationale behind their decisions. Consequently, the latest NLP algorithms have been called \emph{black-box} algorithms.
    Understanding the model's reasoning is essential in various text classification contexts~\citep{ribeiro_why_2016} (e.g., hate speech detection). However, this task is hindered by the black-box nature of these models. Moreover, comprehending the model's reasoning can help establish trust and make informed decisions based on the underlying justifications.

    \definecolor{is}{RGB}{228, 255, 221}
    \definecolor{such}{RGB}{255, 220, 215}
    \definecolor{great}{RGB}{21, 222, 13}
    \definecolor{movie}{RGB}{172, 255, 166}

    \definecolor{This_2}{RGB}{255, 158, 158}
    \definecolor{is_2}{RGB}{21, 222, 13}
    \definecolor{such_2}{RGB}{172, 255, 166}
    \definecolor{great_2}{RGB}{255, 198, 189}
    \definecolor{movie_2}{RGB}{255, 158, 158}

    \begin{figure}[t!]
        \begin{itemize}[noitemsep]
            \item[(a)] {\centering \colorbox{white}{This} \colorbox{is}{is} \colorbox{such}{such} \colorbox{white}{a} \colorbox{great}{great} \colorbox{movie}{movie}\colorbox{white}{!}\par}
            \item[(b)] {\centering \colorbox{This_2}{This} \colorbox{is_2}{is} \colorbox{such_2}{such} \colorbox{white}{a} \colorbox{great_2}{great} \colorbox{movie_2}{movie}\colorbox{white}{!}\par}
        \end{itemize}
        \vspace{-0.5cm}        
        \caption{Examples of local saliency post-hoc explanations from a hypothetical text classifier for a positive movie review. Explanation (a) is more \textit{plausible} than (b). Green means a positive contribution to the model's prediction, and red is negative.}
        \label{fig:explanation_example}
    \end{figure}

    Researchers have developed popular text classification explainability techniques, such as post-hoc local saliency (or heatmap) methods~\citep{tjoa_quantifying_2022, deyoung_eraser_2020}.
    These methods generate heatmaps over tokens (paragraphs, sentences, words, sub-words, or characters) to indicate their significance in the final decision~\citep{ribeiro_why_2016, lundberg_unified_2017, chefer_transformer_2021}\,---\,although their suitability is criticized \cite{bilodeau_impossibility_2024}, these methods are still widely applied \cite{kumari_unintended_2024}.
    The estimation of importance is performed after the decision has been made using an already trained model (i.e., it is post-hoc).
    For instance, \autoref{fig:explanation_example} illustrates word-level saliency explanations that justify the predictions of two trained models in determining whether a movie review is positive or negative.
    In explanation (a), highlighted in green, the most relevant words align well with human expectations, making it intuitive.
    However, in explanation (b), the highlighted words are irrelevant from a human perspective.
    Both explanations may accurately reflect the models' reasoning (thus, they may be \emph{faithful}, according to \citealp{deyoung_eraser_2020}). Nevertheless, they differ in \emph{plausibility}, which refers to the extent to which the explanation matches human intuition~\citep{deyoung_eraser_2020} or is ``convincing of the model prediction'' \citep{jacovi_aligning_2021}.
    
    Ideally, we should be able to enhance the plausibility of a ``non-plausible'' model by ``teaching'' it to provide more plausible explanations. Previous works, such as those by \citealp{strout_human_2019, ross_right_2017, arous_marta_2021, du_learning_2019, mathew_hatexplain_2021}, have explored this concept.
    The reason is that someone training the model clearly understands what a valid explanation should entail.
    However, achieving plausibility while preserving \emph{faithfulness} may require modifying the reasoning of the original model, which in turn risks impacting its performance on the test data.
    Hence, an inherent trade-off exists between model performance and explanation plausibility~\citep{zhang_explain_2021,plumb_regularizing_2020}.

    This paper introduces a methodology that enhances the plausibility of explanations while remaining agnostic to the model architecture and explainability method.
    Our approach incorporates human explanations, represented as \emph{rationales} (i.e., text annotations serving as ground truth for explanations), into text classification models using a novel contrastive-inspired loss. 
    We address the trade-off between classification and the new loss within a multi-objective framework, enabling exploration of the balance between performance and plausibility.
    Unlike other approaches, our methodology does not require modifying the model architecture (e.g., through the addition of attention mechanisms; \citealp{strout_human_2019}) or assuming a specific type of explanation function (e.g., a differentiable explanation function; \citealp{rieger_interpretations_2020}) to incorporate the explanations.
    
    In summary, our contributions are:
    \begin{enumerate}[label=(\roman*)]
        \item A proposal of a novel contrastive-inspired loss function that effectively incorporates rationales into the learning process.
        \item A multi-objective framework that automatically assigns weights to the learning loss and contrastive rationale loss, offering multiple trade-off options between performance and explanation plausibility.
        \item A series of experiments using various models, datasets, and explainability methods, demonstrating the significant enhancement of model explanations without compromising (and sometimes without any detriment to) the model's performance. Notably, our approach exhibits particularly improved plausibility for samples with incorrect explanations.
    \end{enumerate}
    
    We compare our methodology with a previous method from the literature, reinforcing our results. Furthermore, we address the social and ethical implications of ``teaching'' explanations to text classification models. We argue that these concerns are mitigated when the explanations remain faithful to the model's decision-making process.

%% file: content/related_work.tex
\section{Related Work}
    \label{sec:related_work}

    Our work draws on prior research in the areas of rationale utilization and the trade-off between performance and explainability.

    \paragraph{Use of Rationales.}
        Using human annotations to assist machine learning is not a novel concept, as prior works have shown \citep{zaidan_using_2007,zaidan_machine_2008}. 
        Nevertheless, there has been a recent surge in interest in machine learning explainability and fairness, leading to an increased focus on collecting and applying such rationales. 
        Some studies have leveraged rationales to enhance model fairness \citep{rieger_interpretations_2020, liu_incorporating_2019}, while others have explored techniques to extract \citep{zhang_explain_2021, lakhotia_fid-ex_2021, pruthi_weakly-_2020, sharma_computational_2020} or generate \citep{rajani_explain_2019, liu_towards_2019, camburu_e-snli_2018, kumar_nile_2020} model explanations. 
        The most prevalent application of rationales lies in performance improvement, where annotations serve as valuable assistants during the learning process, particularly in tasks involving textual data \citep{sharma_learning_2018, bao_deriving_2018, liu_towards_2019, rieger_interpretations_2020, zhang_explain_2021, arous_marta_2021, mathew_hatexplain_2021, carton_what_2022,ghaeini_saliency_2019, huang_exploring_2021}, images \citep{simpson_gradmask_2019, rieger_interpretations_2020, mitsuhara_embedding_2021}, or tabular data \citep{belem_weakly_2021}.
        In this work, our focus revolves around the incorporation of rationales during model training to ``teach'' explanations, drawing inspiration from the findings of \citet{arous_marta_2021, du_learning_2019, mitsuhara_embedding_2021}.
        In particular, \citet{mathew_hatexplain_2021} collect and annotate a dataset called HateXplain and use its annotations to train a model. Moreover, the UNIREX framework~\cite{chan_unirex_2022} extends this approach to a more general setting.
        
        Importantly, our approach refrains from altering/assuming the model architecture (e.g., by using another model for rationale extraction \citep{chan_unirex_2022}, assuming a model architecture \citep{mathew_hatexplain_2021}, or adding another layer \citep{strout_human_2019, chen_learning_2020, liu_improve_2022, sekhon_improving_2023}) or assuming a specific type of explanation function (e.g., by using input gradients; \citealp{ross_right_2017,ghaeini_saliency_2019}).
        Such interventions are debatable (see \autoref{sec:discussion}) and not always possible.
        Instead, we adopt a model- and explainer-agnostic approach, using rationales to enhance the plausibility of explanations.
        Noticeably, our approach also differs from previous work that rationalizes the input but does not leverage human annotations \cite{lei_rationalizing_2016, bastings_interpretable_2019, jain_learning_2020}.

    \paragraph{Performance and Explainability Trade-off.}
        The existence of a trade-off between machine learning performance and interpretability/explainability is widely debated in the field.
        Several studies have discussed this trade-off \citep{camburu_e-snli_2018, swanson_rationalizing_2020,dubey_scalable_2022,plumb_regularizing_2020,radenovic_neural_2022}. 
        However, differing opinions exist on whether this trade-off always holds, both from a theoretical perspective \citep{jacovi_aligning_2021, rudin_stop_2019} and a practical standpoint \citep{hase_leakage-adjusted_2020}.
        Furthermore, some studies have empirically examined or explored this trade-off \cite{zhang_explain_2021,goethals_non-linear_2022, naylor_quantifying_2021,paranjape_information_2020,jin_simultaneous_2006}.
        Our work shares similarities with the study conducted by \citet{belem_weakly_2021}, as we aim to employ two distinct learning strategies and investigate their trade-offs. 
        However, our approach utilizes different learning strategies, and we conduct the trade-off exploration using a multi-objective optimization algorithm.

%% file: content/theoretical_background.tex
\section{Theoretical Background}
    \label{sec:theoretical_background}

    We define crucial explainability and multi-objective optimization concepts to facilitate a global understanding of our research. We also point to an overview of contrastive learning in \autoref{appendix:contrastive_learning}.

    \subsection{Explainability}
        \label{sec:explainability}

        \paragraph{Rationale.}
        
            In the context of text classification, a \textit{rationale} refers to a snippet extracted from a source text that supports a specific category \citep{deyoung_eraser_2020,carton_what_2022,mathew_hatexplain_2021}. 
            Typically, these rationales are annotated by humans and serve as ground truth explanations for the corresponding categories. 
            For instance, in \autoref{fig:explanation_example}, a typical rationale for the positive class would be ``great movie.''

        \paragraph{Explanation Plausibility.}
        
            The \emph{plausibility} of a model explanation refers to the extent to which it aligns with human intuition \citep{deyoung_eraser_2020} or is considered ``convincing of the model prediction'' \citep{jacovi_aligning_2021}. 
            In practice, this plausibility can be measured by evaluating the agreement between the explanation and the ground truth rationale \citep{deyoung_eraser_2020, jacovi_aligning_2021}.
            Please refer to \autoref{sec:discussion} for a detailed discussion on the pursuit of plausibility.

        \paragraph{Explanation Faithfulness.}
        
            Another crucial aspect of an explanation is its \emph{faithfulness}, which reflects the degree to which the model relies on the explanation to make its prediction \cite{deyoung_eraser_2020}. 
            Following the approach of \citet{deyoung_eraser_2020}, we employ the metrics of \emph{comprehensiveness} and \emph{sufficiency} to quantify faithfulness. 
            We multiply sufficiency by $-1$ to indicate that a higher value is desirable for both metrics.

    \subsection{Multi-objective Optimization}
        \label{sec:multi_objective_optimization}
        
        We aim to investigate the trade-off between model performance and explanation plausibility. \autoref{sec:trade_off} addresses this trade-off exploration by concurrently optimizing two distinct loss functions that may have conflicting objectives. 
        We adopt the definitions that \citet{raimundo_extension_2020} provided for the following concepts.

        \begin{definition}[Multi-objective optimization problem]
            \label{def:moo}
            A \textit{multi-objective optimization problem} (MOO) is an optimization problem with more than one objective, i.e., a problem of the form
            \begin{equation*}
                \begin{split}
                    \min_x& \ \ \ f(x) = (f_1(x), \cdots, f_m(x)),\\
                    \text{subject to}& \ \ \ x \in \Omega \subseteq \mathbb{R}^n, \ \ f \colon \Omega \to \mathbb{R}^m, \ \ f(\Omega) = \Psi.
                \end{split}
            \end{equation*}
        \end{definition}

        Consider two solutions $x_1, x_2 \in \mathbb{R}^n$ where  $f_1(x_1) < f_1(x_2)$ and $f_2(x_1) > f_2(x_2)$. 
        In this case, no clear optimal solution exists. To address this, we introduce the concept of \emph{Pareto-optimality}.

        \begin{definition}[Pareto-optimality]
            A solution $x^* \in \Omega$ is \textit{Pareto-optimal} if there is no other solution $x \in \Omega$ such that $f_i(x) \le f_i(x^*)$ for all $i$ and $f_i(x) < f_i(x^*)$ for some $i$.
        \end{definition}

        The Pareto-frontier comprises objective function values resulting from Pareto-optimal solutions.
        Without considering additional criteria, there is no definitive best solution among them. 
        The decision-maker holds the responsibility of selecting the desired solution.
        While solving a MOO problem poses challenges, various approaches are available.
        Refer to \autoref{sec:moo_theorems} for an overview of the \emph{weighted sum method} and their theoretical foundations.

%% file: content/methodology.tex
\section{Methodology}
    \label{sec:methodology}

    We focus on text classification models to enhance the quality of local saliency post-hoc explanations regarding \emph{plausibility}. 
    We aim to align these explanations with human intuition while maintaining \emph{faithfulness}. 
    To achieve this, we leverage \emph{rationales} to enhance the explanation quality and evaluate the improvement by comparing them with the model explanations.

    \subsection{Notation Description}
        \label{sec:notation_description}

        Consider a multi-class text classification task with classes $C$ and a multi-class text classification model $f_\theta \colon \mathbb{R}^d \to \Delta$. The model takes a text $x \in \mathbb{R}^d$ and produces a probability vector $f_\theta(x) \in \Delta$, indicating the probabilities of $x$ belonging to each class, with parameters $\theta$. 
        Examples of $x$ include TF-IDF vectors \citep{leskovec_mining_2020}, BERT feature vectors \citep{devlin_bert_2019}, or word presence vectors (e.g., Transformer's ``input id'' array; \citealp{vaswani_attention_2017}). We view $f_\theta$ as a black box without assuming any specific structure.
        Let us introduce the explanation function\footnote{$d$ refers to the dimension of the text vector space (e.g., BERT's 768), and $p$ is the number of tokens of a sample.} $e_{f_\theta, k} \colon \mathbb{R}^d \to \mathbb{R}^p$, which assigns a score to each token in $x$, representing its contribution to the $f_\theta(x)$ prediction for class $k \in C$, i.e., $f_\theta(x)_k$. 
        We also have ground-truth human annotations (\textit{rationale}) as a binary vector $e_{x, k} \in \{0, 1\}^p$, indicating the essential tokens for $x$ to be classified as class $k$. 
        The measure of agreement $m \colon \mathbb{R}^p \times \{0, 1\}^p \to \mathbb{R}$ between $e_{f_\theta, k}(x)$ and $e_{x, k}$ quantifies the quality of explanations extracted from $f_\theta$ compared to canonical explanations, reflecting their plausibility.
        Given a set $X = \{X_1, \cdots, X_N\}$ of training texts and a set $y = \{y_1, \cdots, y_N\}$ of training class labels, the commonly used cross-entropy loss is employed during training, defined as:
        \begin{equation}
            \label{eq:loss}
            \begin{split}
                \mathcal{L}_\theta(X, y) =
                -\frac{1}{N} \sum_{i=1}^N \sum_{k=1}^{|C|} \mathds{1}_{y_i = k} \ln \frac{e^{g_\theta(X_i)_k}}{\sum_{j=1}^{|C|} e^{g_\theta(X_i)_j}},
            \end{split}
        \end{equation}
        where, $g_\theta$ represents the logits (pre-softmax) obtained from  $f_\theta$, and $f$ corresponds to the softmax function applied to $g_\theta$. 
        It is worth noting that $\theta$ can represent the training weights of a linear function (in the case of multinomial logistic regression) or a more complex function, such as a neural network.
        
    \subsection{Contrastive Rationale Loss}
        \label{sec:contrastive_loss}

        To enhance the plausibility of model explanations, we incorporate rationales into the model training process. 
        Unlike previous approaches \citep{rieger_interpretations_2020, du_learning_2019,ross_right_2017}, we do not utilize an explanation-based function in the loss function to compare model explanations with ground truth explanations.
        Instead, we construct a loss function for training the text classification model using a modified dataset $\dot X = \{\dot X_1, \cdots, \dot X_N \}$. 
        During training, we replace the full-text $X_i \in \mathbb{R}^d$ with the rationale text $\dot X_i \in \mathbb{R}^d$. 
        By exclusively teaching the model with rationales, we expect them to become the primary basis for the model's decision-making process, leading to correspondingly reflected model explanations\footnote{In this formulation, we assume the explanation function is perfectly faithful, i.e., the explanation results genuinely reflect the model's reasoning. Such a function is not apparent; however, our experimental results suggest that the explainability methods we have access to are sufficient.}.

        In a more general context, $\dot X$ may encompass rationales from a subset or superset of texts in $X$, or even both. 
        In this scenario, $\dot y$ denotes the labels of $\dot X$. 
        Drawing inspiration from the contrastive learning domain \cite{chen_simple_2020,khosla_supervised_2020}, we introduce a novel auxiliary loss function known as the \emph{contrastive rationale loss}:
        \begin{equation}
            \label{eq:contrastive_rationale_loss}
            \begin{split}
                \mathcal{\dot L}_\theta(\dot X, \dot y) =
                -\frac{1}{N} \sum_{i=1}^N \sum_{k=1}^{|C|} \mathds{1}_{\dot y_i = k} \ln \frac{e^{g_\theta(\dot X_i)_k}}{\sum_{j=1}^m e^{g_\theta(\tilde X_{i, j})_k}},
            \end{split}
        \end{equation}
        where $\{\tilde X_{i, j}\}_{j=1}^m$ is a set of $m$ \textit{sample rationales} of $X_i$, i.e., rationales that may be or may be not a ground truth explanation for $X_i$.
        For instance, this set includes the ground truth explanation $\dot X_i$ and other $m - 1$ random rationales, which we call \textit{negative rationales}\,---\,random tokens of $X_i$ uniformly sampled.
        The numerator seeks to maximize the model's output for the rationale in the correct class. At the same time, the denominator aims to minimize the model's output for the random (negative) rationales in the same class.
        Notice that we do not include the explanation function $e_{f_\theta, k}$ (\autoref{sec:notation_description}) in \autoref{eq:contrastive_rationale_loss}, contrary to previous work (\autoref{sec:related_work}).
        This is because we do not want to ``train the explainer'' or ``teach the model how to tweak the explainer.''
        For an in-depth discussion, see \autoref{sec:discussion}.

        The contrastive rationale loss constitutes a particular case when the classifier is a multinomial logistic regression. Further details can be found in \autoref{appendix:logistic_regression}.

    \subsection{Trade-off Exploration}
        \label{sec:trade_off}
        
        \autoref{sec:contrastive_loss} proposes an auxiliary \emph{contrastive rationale loss} function $\mathcal{\dot L}_\theta$  to incorporate rationales during model training. 
        The simultaneous optimization of both cross-entropy $\mathcal{L}_\theta$ and $\mathcal{\dot L}_\theta$ gives rise to a \emph{multi-objective optimization} (MOO) problem (see \autoref{sec:multi_objective_optimization}).
        It is important to note that optimizing both objectives without a trade-off is not feasible. We leverage existing MOO algorithms to explore the trade-off between model performance and explanation plausibility \citep{cohon_multiobjective_1978}.

        In simple terms, MOO solvers such as NISE \citep{cohon_multiobjective_1978}, employing the weighted sum method (\autoref{sec:moo_theorems}), enable trade-off exploration by incorporating hyperparameters $w_1$ and $w_2$ (both $\ge 0$) with $w_1 + w_2 = 1$, and solving the uni-objective problem:
        \begin{equation*}
            \label{eq:trade_off}
            \mathcal{L}_\theta(X, y, \dot X, \dot y) = w_1 \cdot \mathcal{L}_\theta(X, y) + w_2 \cdot \mathcal{\dot L}_\theta(\dot X, \dot y).
        \end{equation*}
        Intuitively, the weight vector $\mathbf{w} = [w_1, w_2]$ controls the trade-off between model performance (original cross-entropy loss) and explanation plausibility (contrastive rationale loss). 
        Increasing $w_2$ from 0 to a positive value explicitly assigns more weight to the contrastive rationale loss. This indicates that the model is trained on data $(\dot X, \dot y)$ that differs from the underlying distribution of $(X, y)$. 
        Consequently, the model's performance on test data, which follows the same distribution as $(X, y)$, is expected to decline. 
        However, since we fit the model using rationales, we alter the model's reasoning, emphasizing the significance of positive rationales within the texts. This emphasis should be reflected in the explanations, as argued in \autoref{sec:contrastive_loss} and demonstrated in our experiments.
        
        MOO solvers like NISE effectively sample representative sets $W_1$ and $W_2$ of trade-off parameters $w_1$ and $w_2$. 
        From the loss optimization process (e.g., \texttt{lbfgs}, SGD, Adam, etc.), these sets yield a set of model weights $\Theta$, where each $\theta \in \Theta$ corresponds to a different classifier $f_\theta \in F_\Theta$. 
        Finally, by searching within the set $F_\Theta$, we can identify Pareto-optimal models that exhibit both performance and plausibility.

%% file: content/experiments.tex
\section{Experiments}
    \label{sec:experiments}

    This section describes experiments to test the methodology proposed in \autoref{sec:methodology}, employing diverse models, datasets, and explainability techniques.
    We aim to verify the usefulness of the contrastive rationale loss (\autoref{sec:contrastive_loss}) in incorporating human rationales and the effectiveness of the MOO solver (\autoref{sec:trade_off}) in finding models that well-represent the Pareto-frontier. Furthermore, we also compare our methodology with previous work.
    Implementation and execution information can be found in \autoref{appendix:implementation}.

    \subsection{Models}
        \label{sec:models}
    
        To evaluate the effectiveness of our method, we assess two types of models: language models and classic NLP models.

        \paragraph{DistilBERT and BERT-Mini.}

            As language model representatives, we test DistilBERT \citep{sanh_distilbert_2020} and BERT-Mini~\citep{turc_well-read_2019}, lightweight versions of the popular BERT \citep{devlin_bert_2019}.
            For fine-tuning on the HateXplain dataset, refer to \autoref{appendix:distilbert}.
            Refer to \autoref{appendix:additional_results} for an additional analysis with BERT-Large.

        \paragraph{TF-IDF with Logistic Regression.}
        
            For classical models, we train a multinomial logistic regression model using TF-IDF vectors \citep{leskovec_mining_2020} (unigrams) with dimensionality reduction to 200 achieved through Truncated Singular Value Decomposition \citep{manning_introduction_2008}. 
        
    \subsection{Datasets and Data Preprocessing}
        \label{sec:datasets}

        \paragraph{HateXplain.}

            This dataset contains annotated hate speech detection samples with human-annotated rationales \citep{mathew_hatexplain_2021}.
            It consists of three classes: normal (without rationales), offensive, and hate speech.
            To address the confounding correlation between offensive and hate speech classes and their rationales, we simplify the dataset by excluding the offensive class (\texttt{hatexplain} dataset).
            We also explore a version including all labels (\texttt{hatexplain\_all} dataset).
            Hereafter, ``HateXplain'' refers to \texttt{hatexplain} unless specified otherwise.

        \paragraph{Twitter Sentiment Extraction (TSE).}

            The TSE \citep{maggie_tweet_2020} is a sentiment analysis dataset containing positive, negative, and neutral tweets with human-annotated rationales. 
            Since neutral class lacks rationales\footnote{TSE neutral class rationales exist but are uninformative because they are the whole sample text in most cases.}, we simplify the classification, excluding this class (\texttt{tse} dataset). 
            An alternative version includes all labels (\texttt{tse\_all} dataset). 
            Hereafter, ``TSE'' refers to \texttt{tse} unless specified otherwise.
            
        \paragraph{Movie Reviews.}

            This dataset comprises positive and negative movie reviews with rationales annotated by humans to support classification \citep{zaidan_using_2007}.

    \subsection{Explainability Methods}
        \label{sec:explainers}

        We utilize two well-known explainers for generating continuous salient maps in textual datasets.

        \paragraph{LIME.}
        
            Short for \textit{Local Interpretable Model-agnostic Explanations} \citep{ribeiro_why_2016}, it creates post-hoc explanations by randomly removing tokens from the text sample and locally approximating the original model predictions using a simpler, interpretable model, which is used to explain the sample's prediction.

        \paragraph{SHAP.}
        
            \textit{SHapley Additive exPlanations} \citep{lundberg_unified_2017} is a model-agnostic explainer that employs Shapley values to explain model predictions.

    \subsection{Explainability Metrics}
        \label{sec:metrics}

        \paragraph{Plausibility.}
            
            We employ the \emph{Area Under the Precision-Recall Curve (AUPRC)} metric to assess the plausibility of model explanations generated by LIME and SHAP. 
            This metric is constructed by varying the threshold over continuous token scores and calculating precision and recall at the token level \citep{deyoung_eraser_2020}.

        \paragraph{Faithfulness.}
            
            We require discrete explanations to evaluate \emph{comprehensiveness} and \emph{sufficiency} (as described in \autoref{sec:explainability}). To address this, we consider the top 1, 5, 10, 20, and 50\% of tokens and average the results, which we refer to as the \emph{Area Over the Perturbation Curve (AOPC)} \citep{deyoung_eraser_2020}.

    \subsection{DistilBERT and HateXplain}
        \label{sec:distilbert_and_hatexplain}

        In this section, we present experimental results to tackle the following research questions: \textit{Does the proposed loss improve explanation plausibility without affecting the performance? Does the MOO solver effectively assist in finding a model with better explanations?}
        We first present a case study with the DistilBERT model and HateXplain dataset to showcase the main results of our experiments.
        \autoref{sec:other_experiments} shows other results.
        The explainability metrics (plausibility and faithfulness) are computed only for the hate speech class because the normal class lacks rationales.

        The DistilBERT model trained only with cross-entropy loss achieves a test accuracy of 84.8\% with balanced recall among classes.
        \autoref{fig:bad_explanations}~(a) illustrates an example of a bad explanation extracted from this model. It shows that even high-performing classifiers can also present unreasonable explanations.

        \begin{figure}[t!]
            \begin{itemize}[noitemsep]
                \item[(a)] {\centering \colorbox{white}{ugh} \colorbox{white}{i} \colorbox{white}{hate} \colorbox{white}{d*kes} \colorbox{movie}{😐}\par}
                \item[(b)] {\centering \colorbox{white}{ugh} \colorbox{white}{i} \colorbox{white}{hate} \colorbox{movie}{d*kes} \colorbox{white}{😐}\par}
            \end{itemize}
            \vspace{-0.5cm}
            \caption{Examples of explanations of the hate speech class. Explanation (a) is from the original model, and (b) is from the model with top-AUPRC. Green means a positive contribution to the model's prediction. The top-1 token was selected for visualization purposes. More examples in \autoref{tab:more_bad_explanations}.}
            \label{fig:bad_explanations}
        \end{figure}
        
        \begin{figure}[ht]
            \centering            
            \includegraphics[width=0.89\linewidth]{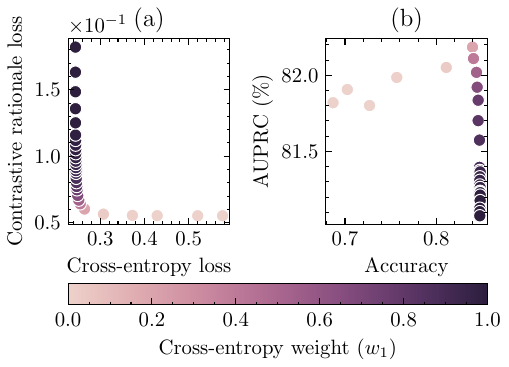}            
            \caption{(a) Trade-off between the two losses on the training data. (b) Trade-off between accuracy and plausibility of the test data. The color scale represents the cross-entropy weight $w_1$ (\autoref{sec:trade_off}). We ignore the model with $w_1 = 0$ as it is out of scale. Results including $w_1 = 0$ and shared scale between axes are in \autoref{appendix:additional_results}.}
            \label{fig:accuracy_auprc}            
        \end{figure}

        We employ NISE \citep{cohon_multiobjective_1978} to find 30 models that well-represent the Pareto-frontier using the cross-entropy and the contrastive rationale loss (using 2 random, negative rationales) on the training data. \autoref{fig:accuracy_auprc}~(a) reveals that the two losses are conflicting, particularly for non-extreme values of $w_1$.
        
        For each model in the frontier, we evaluate the model's performance and the explanation plausibility on the test data (\autoref{fig:accuracy_auprc}~(b)).
        Plausibility was measured using mean AUPRC, comparing LIME's explanations with ground truth rationales.
        \autoref{fig:accuracy_auprc}~(b) shows that, as NISE increases the weight of the contrastive rationale loss during training, the plausibility increases almost without hurting performance: the top-plausibility model had a relative
        increase of 1.4\% in AUPRC (an absolute increase of 1.1\%), despite a relative decrease of 0.9\% in accuracy (an absolute decrease of 0.8\%).
        At some point, performance and explanation quality deteriorate, given that the training without the cross-entropy is meaningless.
        We noticed that around 51\% of the best-explained samples originally had AUPRC equal to 1.
        By disregarding these samples, the AUPRC relative increase becomes 5.3\% (absolute increase becomes 3.3\%). At the same time, the high AUPRC explanations have a relative and absolute decrease of less than 1\%  (\autoref{fig:sample_level_auprc}).
        The inadequate explanations are being improved without significantly harming the good explanations (see example in \autoref{fig:bad_explanations}; more examples in \autoref{tab:more_bad_explanations}).

\begin{figure}[ht]
            \includegraphics[width=0.9\linewidth]{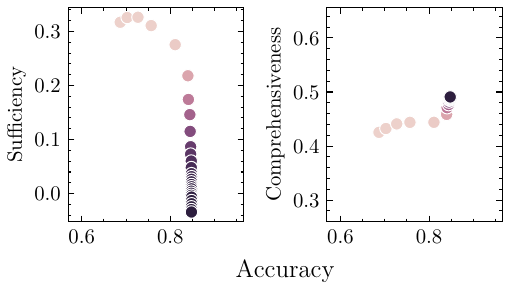}
            \caption{Trade-off between accuracy and faithfulness (sufficiency and comprehensiveness) on test data. 
            Higher values are better.
            The color scale is the same as the previous figures. The data scale is equal between the two graphics and their x- and y-axes.}
            \label{fig:accuracy_faithfulness}
        \end{figure}

        Finally, we must guarantee faithful explanations (i.e., they genuinely represent the models' reasoning) when we strengthen the training with rationales.
        \autoref{fig:accuracy_faithfulness} presents the trade-off between performance and explanation faithfulness on test data.
        Sufficiency tends to increase as we strengthen the training with rationales, while comprehensiveness tends to decrease. 
        However, the explanations are becoming more sufficient without significantly losing comprehensiveness (sufficiency's variation is an order of magnitude higher than the comprehensiveness').

        In summary, the results present a desirable scenario in which \textit{one trades-off a small decrease in accuracy for a reasonable increase in explainability quality (both plausibility and sufficiency)}, especially for originally bad explanations. \textit{The MOO solver effectively assists in finding a model with better explanations.}

    \subsection{Experiments With All Models and Datasets}
        \label{sec:other_experiments}

\begin{figure}[t!]
            \centering
            \includegraphics[width=\linewidth]{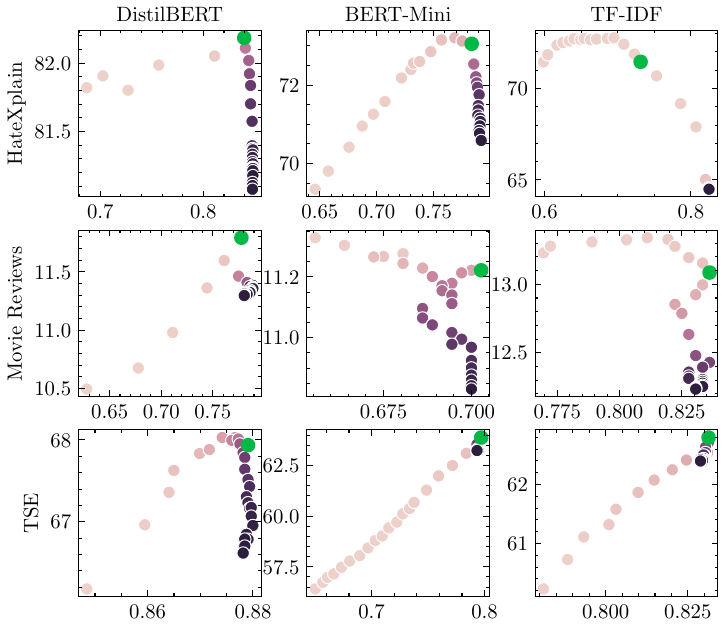}
            \caption{Trade-offs between performance (accuracy, x-axis) and plausibility (AUPRC, y-axis, in percentage (\%)) for all models and datasets (test data). There are 2 random (negative) rationales, and the explainer is LIME. Green dots are the models chosen to be analyzed more carefully. The color scale is the same as the previous figures. We ignore the model with $w_1 = 0$ in all graphics as it is out of scale. Larger figure and results including $w_1 = 0$, 5 rationales and/or SHAP, shared scale between axes, and Pareto-frontiers are in \autoref{appendix:additional_results}.}
            \label{fig:complete_graphic}
        \end{figure}

        Now, we evaluate our framework in all models, datasets, and explainability techniques that we consider in this paper.
        Specifically, we aim to discover \textit{whether the previous results (usefulness of the contrastive loss and effectiveness of the MOO solver) extend to the general case.}
        \autoref{fig:complete_graphic} overviews all performance vs. plausibility trade-offs on test data.
        The number of random (negative) rationales used is 2, and the explainer is LIME.
        To comprehend its effect, we also test with 5 rationales and/or explainer SHAP (\autoref{appendix:additional_results}).
        \autoref{fig:complete_graphic} shows a non-constant shape of the final frontier across all experiments.
        For instance, while TF-IDF trades accuracy for plausibility in the HateXplain dataset, it increased both dimensions in TSE.
        However, the shape is the same when changing the number of negative rationales (\autoref{fig:complete_graphic_lime_5}) and similar when the explainer is SHAP (Figures \ref{fig:complete_graphic_shap_2} and \ref{fig:complete_graphic_shap_5}).
        Finally, despite the TSE dataset having a higher number of poor-performing models, the improvement for a well-selected model is not negligible (\autoref{tab:complete_graphic}).

        \begin{table*}[ht]
            \centering
            \caption{Comparison between the original model (cross-entropy only) and the chosen model (green dots on \autoref{fig:complete_graphic}) for each performance and explainability metric on test data. 
            ``rel.'' means relative variation. The column $w_1$ indicates the weight $w_1$ of the chosen model's cross-entropy loss during training. Number of negative rationales is 2, and the explainer is LIME. A complete table (with 5 negative rationales and/or SHAP) is available in \autoref{appendix:additional_results}.}
            
            \label{tab:complete_graphic}
            \begin{tabular}{ccc|ccccc}
                \hline
                \textbf{Dataset}           & \textbf{Model} & \textbf{$w_1$} & \textbf{Acc. \%} & \textbf{AUPRC \%} & \textbf{AUPRC rel. \%} & \textbf{Suff.} & \textbf{Comp.} \\
                \hline
                \multirow{3}{*}{HateXplain} & DistilBERT    &   0.20 &  -0.80 &  1.11 &         1.37 &   0.25 &  -0.03 \\
                    & BERT-Mini     &   0.29 &  -0.84 &  2.46 &         3.49 &   0.40 &  -0.05 \\
                    & TF-IDF        &  0.002 &  -9.35 &  6.96 &        10.79 &   0.13 &  -0.10 \\
                \hline
                \multirow{3}{*}{Movie Reviews} & DistilBERT &   0.12 &  -0.28 &  0.50 &         4.39 &   0.25 &  -0.05 \\
                    & BERT-Mini  &   0.26 &   0.28 &  0.39 &         3.61 &   0.00 &  -0.02 \\
                    & TF-IDF     &   0.09 &   0.56 &  0.85 &         6.95 &  0.00 &   0.01 \\
                \hline
                \multirow{3}{*}{TSE} & DistilBERT           &   0.64 &   0.09 &  1.32 &         1.98 &   0.05 &  0.00 \\
                    & BERT-Mini            &   0.19 &   0.37 &  0.64 &         1.01 &   0.06 &   0.01 \\
                    & TF-IDF               &   0.42 &   0.24 &  0.40 &         0.64 &   0.01 &  -0.02 \\
                \hline
            \end{tabular}
        \end{table*}

        The green dots in \autoref{fig:complete_graphic} represent the models manually selected as ``good choices'' of the trade-off between performance and plausibility. We analyzed them more carefully and compared them to the original models (i.e., $w_1 = 1$, darkest point on the figures).
        For example, the green dot of DistilBERT with HateXplain is an obvious choice because it improves AUPRC without harming performance.
        Conversely, TF-IDF with HateXplain trades one metric for the other. Thus, a few dots were chosen with some degree of ``good judgment.''
        \autoref{tab:complete_graphic} compares the original and selected models.
        All models improved the plausibility of their explanations, in some cases marginally (as for the TSE dataset).
        The accuracy generally varies slightly, positive and negative, except for a significant drop of TF-IDF with HateXplain.
        Finally, sufficiency is generally positive, with significant improvements for the language models. At the same time, the comprehensiveness is usually negative but an order of magnitude smaller than the improvements in sufficiency.
        Results for SHAP and 5 negative rationales are in \autoref{tab:complete_graphic_all} and, because the trade-off shapes of Figures \ref{fig:complete_graphic}, \ref{fig:complete_graphic_lime_5}, \ref{fig:complete_graphic_shap_2} and \ref{fig:complete_graphic_shap_5} are similar, they present similar conclusions, showing the robustness of our framework for different explainers and number of rationales.
        For examples of explanation improvement, refer to Tables \ref{tab:more_bad_explanations} and \ref{tab:more_more_bad_explanations}.

        In general, \textit{all models improve their explanation quality in plausibility (and the majority of them in sufficiency, too) without harming the performance significantly}, showing the robustness of our framework.
        \textit{The multi-objective exploration was essential to find the best trade-offs.}
        Conclusions are similar for non-binary classification (see \autoref{appendix:additional_results}).

    \subsection{Methodology Comparison}
        \label{sec:comparison_with_hatexplain}

        In HateXplain's paper \citep{mathew_hatexplain_2021}, the authors test their dataset by proposing BERT-HateXplain, a BERT version incorporating the rationales as an additional input. They incorporate the annotations using a novel loss function over the attention weights of the last layer of BERT\footnote{Their attention loss is multiplied by a ``trade-off'' hyperparameter $\lambda$. We use their suggestion of $\lambda$ values (\autoref{appendix:implementation}).},
        which is a particular case of the UNIREX framework \cite{chan_unirex_2022}.
        We compare our methodology with the BERT-HateXplain model, using the same dataset (\texttt{hatexplain\_all}), model (\texttt{bert-base-uncased}), and explainer (LIME), and setting the number of random (negative) rationales to 2.

        \definecolor{hatexplain}{RGB}{73,151,201}
        \definecolor{ours}{RGB}{135,78,126}

        \begin{figure}[t!]
            \centering                
            \includegraphics[width=0.89\linewidth]{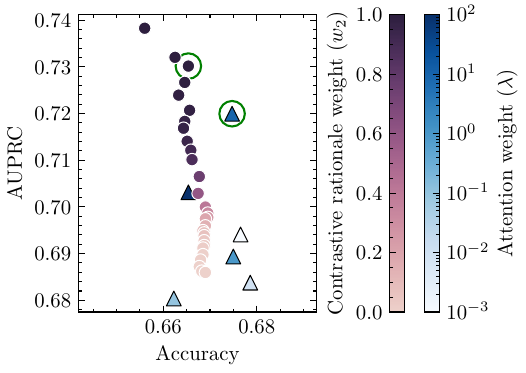}            
            \caption{
                Comparison between BERT-HateXplain (\berthatexplainsquare) and our methodology 
                (\ourscircle) on test data.
                Number of negative rationales is 2 for our method. Color scales indicate the explanation weights $\lambda$ (for HateXplain, log scale) and $w_2$ (for our method).
                As usual, we ignore the model with $w_2 = 1$ as it is out of scale.
                Circled points are the chosen models for each method to be analyzed more carefully.
                Data scale is equal between x- and y-axes.
            }            
            \label{fig:comparison_graphic}
        \end{figure}

        \begin{table}[ht]
            \caption{Comparison between the chosen models (circled points in \autoref{fig:comparison_graphic}) of BERT-HateXplain and our method on test data. Accuracy and AUPRC are in percentage (\%).}
            \label{tab:comparison_graphic}            
            \begin{tabular}{l|cccc}
                \hline
                \textbf{Model} & \textbf{Acc.} & \textbf{AUPRC} & \textbf{Suff.} & \textbf{Comp.} \\

                \hline
                HateXplain &  \textbf{67.47} &   72.00 &  0.12 &  \textbf{0.53} \\
                Ours            &  66.54 &   \textbf{73.02} &  \textbf{0.14} &  0.40 \\
                \hline
            \end{tabular}            
        \end{table}
        
        \autoref{fig:comparison_graphic} presents the trade-off between accuracy and plausibility (mean AUPRC) on test data for BERT-HateXplain and our methodology after optimization on training data.
        For BERT-HateXplain, we use the suggested hyperparameters from their paper \citep{mathew_hatexplain_2021}.
        The shape of our curve is similar to the other experiments involving language models.
        BERT-HateXplain has a less stable curve because their model training is stochastic, while our methodology is deterministic (\autoref{sec:trade_off}).
        The circled dots are the chosen models using a ``good judgment'' of improving AUPRC without hurting too much accuracy. \autoref{tab:comparison_graphic} compares the selected models for each method.
        Our methodology has better plausibility, while BERT-HateXplain has better accuracy.
        Additionally, our methodology has better sufficiency, while BERT-HateXplain has better comprehensiveness.
        These results align with the canonical BERT-HateXplain results \citep{mathew_hatexplain_2021} in their absolute values and conclusion: they improve performance and comprehensiveness while decreasing sufficiency.
        Importantly, our method does not require any assumption of model architecture, while BERT-HateXplain does. This comparison expands the results of the other experiments, showing that our methodology can trade a little of performance to improve explanation quality (by improving plausibility while keeping faithfulness) in a model-agnostic approach.

    \subsection{Further Experiments}

        We performed additional experiments to assess our methodology further (\autoref{appendix:additional_results}). We found that the performance of our method for larger models is similar to other experiments and that we can improve out-of-distribution performance.

%% file: content/discussion.tex
\section{Discussion}
    \label{sec:discussion}

    \paragraph{Should We Model Plausibility?}
    
        \citet{jacovi_aligning_2021} argue that explanation plausibility should not be pursued because it is an ethical issue: the explainer would pursue convincing the user of the model decision, possibly providing unfaithful justifications. Our perspective is different: the explainer is never adjusted to convince the user (the model explainer is not ``trained'' with rationales, and the model does not learn how to tweak the explainer). Instead, we update the model's internal decision, aiming for better explanations. Our perspective is more aligned with \citet{zhou_exsum_2022} who defends that plausibility contributes to \textit{understandability}: ``given the same level of correctness, a higher-alignment explainer may be preferable'' \cite{zhou_exsum_2022}.

        \paragraph{Is There Really a Trade-Off?}

            The hypothesis of this work is the existence of a trade-off between model performance and explanation plausibility.
            This happens because, once we fix the model's architecture, it is impossible to promote more alignment with the rationales without changing its optimal.
            The Pareto frontier in \autoref{fig:complete_graphic_lime_2_frontier} clearly shows that there is not any model that is better than all the others in both metrics (exceptionally for one case), further indicating the presence of a trade-off in its classic sense.
            \autoref{sec:related_work} presents references that argue both in favor and against in the debate of the existence of a trade-off.
            This work contributes to this debate by proposing an explicit trade-off formulation (Equations \ref{eq:loss} and \ref{eq:contrastive_rationale_loss}) and experiments exploring the existence of this trade-off.

        \paragraph{Model and Explainer Agnosticism.}
        
            Our approach claims to be model- and explainer-agnostic because we only influence the training procedure by adding another loss function that incorporates the rationales.
            We do not specify model type \citep{strout_human_2019, mathew_hatexplain_2021} or ask for a specific type of explanation function \citep{rieger_interpretations_2020}.

        \paragraph{Light Hyperparameter Search.}
        
            The trade-off is explored using a MOO solver to identify optimal weights. Model training is confined to the classification layer, akin to training logistic regression in the latent space (see \autoref{appendix:implementation}). Inference across the language model occurs just once. This approach eliminates the need for fine-tuning, rendering the optimization process both convex and expedient.

        \paragraph{Data Distribution Shift.}

            The introduction of rationales, with a decurrent performance drop, can be interpreted as a data distribution shift.
            To limit its effect on the performance, we keep the original classification loss and find the right balance between explanation plausibility and performance drop.

        \paragraph{Other Benefits.}

            To change the shortcuts that neural networks explore to perform tasks, it is necessary to update most, if not all, of the model's weights. Despite our work training weights of the final layer only, we believe that reducing network shortcuts with our method should be explored in future work.
            Training models to have more plausible reasoning can decrease biases, improving users' trust. In future work, we intend to perform a large-scale user trust evaluation.

        \paragraph{Datasets Diversity.}

            We explored a diverse set of datasets used in the literature \cite{mathew_hatexplain_2021, atanasova_diagnostic_2020}. They vary in text and rationale length, text distribution, and number of classes (\autoref{appendix:additional_results}). They include complex and ambiguous rationales (e.g., Movie Reviews) and those with nuanced classification categories, such as the ``offensive'' and ``hatespeech'' classes in HateXplain (\autoref{tab:three_classes_graphic}).

%% file: content/conclusion.tex
\section{Conclusion}
    
    We propose a novel approach for enhancing the explanation plausibility of text classification models by incorporating human rationales, which capture human knowledge. 
    Our method is model-agnostic and explainability method-agnostic, making it compatible with various model architectures and explainers. 
    We introduce a new contrastive-inspired loss function that integrates the rationales into the learning process. 
    We demonstrate the feasibility of finding models that achieve a trade-off between improved plausibility and a minimal or negligible decrease in model performance. 
    A comparative analysis establishes the superior effectiveness of our approach in enhancing plausibility while maintaining faithfulness and model agnosticism. 
    We validate our method using a diverse set of explainers, datasets, and models encompassing modern and traditional NLP models.
    Furthermore, we envision the potential extension of our approach to accommodate other explainers, datasets, and models, offering a seamless pathway to enhancing the plausibility of text classification algorithms.

%% file: content/limitations.tex
\section*{Limitations}

    \myparagraph{Model Agnosticism.}    
        The employed multi-objective optimization (MOO) solver, NISE, demands convex objective functions.
        We claim our method is agnostic to any classification model, and this is true. However, when dealing with models that do not satisfy the convexity condition, e.g., complex neural networks, one should employ other MOO algorithms.
        To circumvent this limitation with the language models, we trained only the classification layer or first fine-tuned the model with cross-entropy loss (\autoref{appendix:implementation}).

    \myparagraph{DistilBERT and BERT-Mini.}    
        DistilBERT and BERT-Mini, as they are Transformer encoder-based models, do not scale to long texts because of the limited input size.
        We did not approach this limitation in this work, and we plan this for future work.
        For our long text dataset, Movie Reviews, we truncated the text to the input size of the model, which may have impacted the results.
    
    \myparagraph{Larger Datasets.}
        To the best of our knowledge, there is a limitation in the literature regarding the availability of large classification textual datasets with human annotations in the sentence/phrase/word/token level \cite{wiegreffe_teach_2021}. Other tasks, such as natural language inference \cite{camburu_e-snli_2018}, are out of the scope of this work. Conducting large dataset annotations is intended for future work.

    \myparagraph{Model Scaling.}
        In our methodology, only the classifier layer is trained, diminishing the benefits of further scaling the underlying model responsible for generating representations. Additionally, computational limitations become a significant factor when evaluating models with explainers, as these methods necessitate thousands of inferences for each sample. Despite these constraints, our experiments with BERT-Large indicate that findings are consistent even with larger models. It is also noteworthy that BERT-based models remain relevant benchmarks in recent language model research, as evidenced by studies such as from \citet{du_generalizing_2023}.
    
    \myparagraph{Annotation Efforts.}
        We are aware of the additional effort required to collect annotations for textual datasets and how this limits the extension of our work's application. However, we notice that, to make models ``learn with humans,'' human efforts must be made to ``teach machines.'' We believe this is a limitation of the problem (``learning with explanations'') instead of our work (a specific methodology to incorporate the explanations).
        Even so, there is a relevant availability of textual datasets with annotations \cite{wiegreffe_teach_2021}. Finally, recent advances in crowdsourcing annotation systems allow an efficient annotation of datasets at scale \cite{drutsa_crowdsourcing_2021}.

    \myparagraph{Human Study.}
        Consistent with precedents in the field \cite{mathew_hatexplain_2021,ross_right_2017}, we did not conduct a separate human evaluation. This decision is based on the redundancy of such an evaluation with the existing human annotations in our dataset. Any human assessment would only assess the machine's rationale against individuals' subjective interpretations of the rationale. This process is equivalent to the annotation process already undertaken.

    \myparagraph{Methodology Comparison.}
        BERT-HateXplain is an appropriate baseline for our approach, sharing the same explanation method, dataset, and metrics. It aptly represents other baseline methods \cite{chan_unirex_2022,zhang_explain_2021,lakhotia_fid-ex_2021,arous_marta_2021,strout_human_2019}, which also integrate rationale extraction in the forward pass and learn from annotated rationales. Future work will include comparisons with gradient saliency-based baselines \cite{ghaeini_saliency_2019, huang_exploring_2021}. Furthermore, BERT-HateXplain is a specific instance of UNIREX \cite{chan_unirex_2022}. The only difference in its ``Share LM'' variant (model and extractor with shared parameters) is an additional faithfulness loss beyond our current scope. The ``Double LM'' variant of UNIREX, featuring a distinct architecture for explanation extraction, is also outside our study's purview.

%% file: content/ethics_statement.tex
\section*{Ethics Statement}

    Some authors consider pursuing plausibility as an ethical issue \citep{jacovi_aligning_2021}.
    Part of this work argues this is not the case (\autoref{sec:discussion}).
    In this work, we utilize a hate speech detection dataset and train models with this data.
    We do not intend to publicly distribute the trained models as they may incorporate strong, toxic biases.

%% file: content/acknowledgements.tex
\section*{Acknowledgements}

    This work was supported by the National Council for Scientific and Technological Development (CNPq) under Grant \#311144/2022-5, Carlos Chagas Filho Foundation for Research Support of Rio de Janeiro State (FAPERJ) under Grant \#E-26/201.424/2021, São Paulo Research Foundation (FAPESP) under Grant \#2021/07012-0, the School of Applied Mathematics at Fundação Getulio Vargas, and FAEPEX-UNICAMP under Grants 2559/22 and 2584/23.
    We also thank Vicente Ordonez and the anonymous reviewers for their important feedback.

%% file: appendix.tex
\clearpage

\section{Multi-objective Optimization Theorems and Definitions}
    \label{sec:moo_theorems}
 
    The \emph{weighted sum method} is an approach to solve a MOO problem.
    It balances the objective functions and converts the problem into a uni-objective form.

    \begin{definition}[Weighted sum method]
        Given a MOO problem as in \autoref{def:moo}, the \textit{weighted sum method} transforms the problem into
        \begin{equation*}
            \begin{split}
                \min_x \ \ \ &w^\intercal f(x),\\
                \text{subject to} \ \ \ &x \in \Omega \subseteq \mathbb{R}^n, f \colon \Omega \to \mathbb{R}^m, f(\Omega) = \Psi,\\
                &\sum_{i=1}^m w_i = 1, w \in \mathbb{R}_+^m.
            \end{split}
        \end{equation*}
    \end{definition}

    With a few assumptions, solving the weighted problem is necessary and sufficient to search for the Pareto-frontier of the original MOO problem.

    \begin{theorem}[Necessity]
        \label{theo:necessity}
        If $w \in (\mathbb{R}_+^{*})^m$ and $x^*$ is a solution of the weighted problem, then $x^*$ is a Pareto-optimal solution of the original MOO problem.
    \end{theorem}
    \begin{proof}
        Following \citet{raimundo_extension_2020}, suppose, by contradiction, that $x^*$ is a solution to the weighted problem (with weights $w$) but not a Pareto-optimal solution.
        Then, there exists $x$ such that, for some $i$, $f_i(x) < f_i(x^*)$ and, for all $j$, $f_j(x) \le f_j(x^*)$, by definition.
        Then there exists $\varepsilon \ge 0$ such that $f(x) + \varepsilon = f(x^*)$, with $\varepsilon_i > 0$.
        Finally, $w^\intercal f(x) + w^\intercal \varepsilon = w^\intercal f(x^*)$, which means $w^\intercal f(x) < w^\intercal f(x^*)$. Absurd.
    \end{proof}

    \begin{theorem}[Sufficiency]
        If the original MOO problem is convex, for any Pareto-optimal solution $x^*$ there exists a weighting vector $w$ such that $x^*$ is the solution of the weighted problem.
    \end{theorem}
    \begin{proof}
        This theorem was proved by \citet[Theorem 3.1.4]{miettinen_nonlinear_1998}.
    \end{proof}

    The equivalence between the MOO problem and the weighted problem, established when the MOO problem is convex, is crucial.
    It enables multi-objective optimization algorithms that characterize the Pareto-frontier using the weighted sum method (e.g., NISE, \citealp{cohon_multiobjective_1978}).

\section{Contrastive Loss for Logistic Regression}
    \label{appendix:logistic_regression}

    The logistic regression as the classifier is a particular case that deserves a highlight.
    When the model $f_\theta$ is a multinomial logistic regression over text embedding vectors, we can represent the contrastive rationale loss function in the following way:
    \begin{equation}
        \label{eq:logistic_regression_contrastive_loss}
        \begin{split}
            &\mathcal{\dot L}_\theta(\dot X, \dot y) = \\
            &-\frac{1}{N} \sum_{i=1}^N \sum_{k=1}^{|C|} \mathds{1}_{\dot y_i = k} \ln \frac{\exp (\dot X_i \cdot \theta_k)}{\sum_{j=1}^m \exp (\tilde X_{i, j} \cdot \theta_k)}.
        \end{split}
    \end{equation}
    The dot product between two vectors is commonly used as a similarity function in a contrastive learning context \citep{khosla_supervised_2020}.
    When minimizing \autoref{eq:logistic_regression_contrastive_loss}, one is training an \textit{anchor} $\theta_k$ to approximate a \textit{positive rationale} $\dot X_i$ and to distance \textit{negative rationales} $\{\tilde X_{i, j}\}_{j=1}^m \setminus \{\dot X_i\}$, just like in contrastive learning.
    However, positive and negative vectors cannot be optimized in our case.

    The multinomial logistic regression as a model is analogous to a neural network with all but the classification layer's weights frozen.
    When there are only two classes, it is easy to prove that binary and multinomial logistic regression are equivalent.
    Finally, the logistic regression results in a loss function $\mathcal{\dot L}$ that is convex with respect to the weights $\theta$, easing the search for the model performance vs. explanation plausibility Pareto-frontier through the employing of convex multi-objective optimization algorithms, e.g., NISE (\citealp{cohon_multiobjective_1978}; \autoref{sec:moo_theorems}).

\section{Contrastive Learning Theoretical Background}   
    \label{appendix:contrastive_learning}

    Consider a scenario where samples belonging to a group $p$ follow the distribution $\mathcal{T}_p$. In contrastive learning, the objective is to ensure that the representations of samples originating from the same distribution, $\{T_{p, i}\}_i \sim \mathcal{T}_p$, exhibit similarity in the vector space while samples from different distributions are positioned further apart. To achieve this, the learning process aims to maximize a chosen agreement metric among vector representations of samples from the same distribution while simultaneously minimizing this agreement for samples from different distributions.

    In visual representations, \citet{chen_simple_2020} employ a contrastive loss function in the latent space to maximize the agreement between two preprocessed versions of the same image while minimizing the agreement between preprocessed versions of different images. 
    Similarly, \citet{khosla_supervised_2020} propose a \emph{supervised contrastive loss} that maximizes the agreement between images belonging to the same class while minimizing the agreement between images from different classes.

\section{DistilBERT and BERT-Mini Fine-tuning on HateXplain}
    \label{appendix:distilbert}

    The rationales of the HateXplain dataset contain words not included in the original \texttt{distilbert-base-uncased}\footnote{Available at \url{https://huggingface.co/distilbert-base-uncased}} and \texttt{bert-mini}\footnote{Available at \url{https://huggingface.co/prajjwal1/bert-mini}} model's vocabulary because they are offensive and hate speech words.
    However, when training a model to incorporate rationales, including these tokens in the vocabulary may be important. Otherwise, the results would be underestimated.
    In the train portion of the dataset, we filtered the most popular out-of-vocabulary tokens (those with more than ten occurrences), added them to the models' vocabularies, and fine-tuned the models in this portion.
    We used a masked language modeling probability of 0.15 with a batch size of 8 for 15 epochs in a GPU NVIDIA GeForce GTX 1070.
    We do not apply this process for the methodology comparison to keep similarities with the original HateXplain work \citep{mathew_hatexplain_2021}.

\section{Implementation and Execution}
    \label{appendix:implementation}

    \paragraph{Logistic Regression.}

        We implemented the Logistic regression with Scikit-learn.
        Its implementation was adapted to incorporate the contrastive rationale loss.
        The experiments used the following hyperparameters: tolerance of 1e-4, max iterations of 1e3, $l2$ penalty, \verb|lbfgs| solver, and \verb|multinomial| implementation.
        The $C$ hyperparameter was chosen with cross-validation on the training set.
        The regularization term is added to the two losses (cross-entropy and contrastive rationale loss).
        Therefore, when the two losses are weighted by $\mathbf{w}$, the regularization term comes with weight 1.

    \paragraph{DistilBERT and BERT-Mini.}
    
        The DistilBERT version used in this work was the \texttt{distilbert-base-uncased}\footnote{Available at \url{https://huggingface.co/distilbert-base-uncased}}, while the BERT-Mini version was the \texttt{prajjwal1/bert-mini}\footnote{Available at \url{https://huggingface.co/prajjwal1/bert-mini}}.
        The models are used for text classification; therefore, we plug a classification head on top of the \texttt{[CLS]} output vector.
        We keep all but the classification layer's weights frozen to guarantee the loss convexity (as we pointed out in \autoref{appendix:logistic_regression}), and the models are easier to train.
        These models were not trained with gradient descent because only a classification layer was trained.
        The classification layer was implemented as a multinomial logistic regression and trained accordingly (see previous paragraph).
        The inference over the DistilBERT and BERT-Mini models was performed using GPUs NVIDIA Quadro RTX 6000 and NVIDIA GeForce GTX 1070. The running time of all experiments took the order of magnitude of a month.
        The models truncate the input text to their input limit length of 512.
        The LIME's disturbed text input has its tokens substituted by \texttt{[MASK]} for these models, keeping the original text sample length.

    \paragraph{Datasets.}

        In the HateXplain dataset, because more than one annotator is used for each sample, we apply majority consensus to both rationale and class assignments, disregarding non-consensual samples.

        The HateXplain dataset is already tokenized, and Movie Reviews was tokenized with Python's \texttt{str.split()}. Tweet Sentiment Extraction (TSE) was tokenized using \texttt{re.split(f"([\textbackslash\textbackslash s\{punctuation\}])", str)} with \texttt{punctuation} imported from \texttt{string} and with regex special characters escaped.
        \autoref{tab:description_datasets} presents a description of the datasets.
    
        \begin{table}[ht]
            \centering
            \caption{Description of the datasets after filtering (\autoref{sec:datasets}). HateXplain average rationale length is calculated over the hate speech class only, and \texttt{hatexplain\_all}, over hate speech and offensive classes.}
            \label{tab:description_datasets}
            \begin{tabular}{cccc}
                \hline
                \textbf{Dataset}                                        & \textbf{Samples} & \textbf{\begin{tabular}[c]{@{}c@{}}Average\\ sample\\ length\end{tabular}} & \textbf{\begin{tabular}[c]{@{}c@{}}Average\\ rationale\\ length\end{tabular}} \\ \hline
                HateXplain                                              & 13749            & 23.9                                                                       & 3.4                                                                           \\ \hline
                \begin{tabular}[c]{@{}c@{}}\texttt{hatexplain}\\ \texttt{\_all}\end{tabular}                                              & 19228            & 23.4                                                                       & 3.3                                                                           \\ \hline
                \begin{tabular}[c]{@{}c@{}}Movie\\ Reviews\end{tabular} & 1800             & 741.7                                                                      & 62.1                                                                          \\ \hline
                TSE                                                     & 16330            & 17.5                                                                       & 4.7                                                                           \\ \hline
                \texttt{tse\_all}                                                     & 27378            & 17.0                                                                       & 9.2                                                                           \\ \hline
            \end{tabular}
        \end{table}

    \paragraph{LIME.}

        The LIME explainer was implemented using 1000 samples, and the number of features was the number of tokens of the text sample.
        It applied the perturbations using each dataset's tokenization and filled the perturbed tokens in accordance with the model requirements.
        For instance, DistilBERT and BERT-Mini required the perturbed tokens to become \verb|[MASK]| tokens to keep the input sequence length unchanged.

    \paragraph{Comparison with HateXplain.}

        To compare our methodology with HateXplain's \citep{mathew_hatexplain_2021}, we implement their model in both their and our framework.
        We tried to keep the implementation, including methods and hyperparameters, as close as possible to the details in their paper \citep{mathew_hatexplain_2021} and in their GitHub repository\footnote{\url{https://github.com/hate-alert/HateXplain}}.
        We use the three-class HateXplain dataset (\texttt{hatexplain\_all}), the model \texttt{bert-base-uncased}, and the explainer LIME.
        In our method, we also use 2 negative (random) rationales.
        In particular, BERT's input length limit is set to 128 tokens. Finally, we use the BERT's \texttt{pooled\_output} vector as input to the classification layer, in contrast to the other language models in this paper, in which we use the \texttt{[CLS]} token output vector.

        In our methodology, before exploring the trade-off between cross-entropy and the contrastive rationale loss using NISE, we fine-tune the model with the cross-entropy loss only.
        This is done to maintain performance compatibility between our method and HateXplain's, which fine-tunes the model to train the attention.
        However, we do not apply the fine-tuning procedure of \autoref{appendix:distilbert}, i.e., incorporating new tokens into the model's vocabulary and training the model in the masked language model task (MLM).
        This could be performed, but it would differ from what was done in HateXplain's work.

        The model's hyperparameters (in their methodology and in our fine-tuning) were set to the following values: learning rate of $2e-5$, attention softmax temperature parameter of 0.2, Adam optimizer, standard BERT dropouts of 0.1, 6 heads of attention supervision in the last BERT layer, batch size of 16, 20 epochs, and epsilon of $1e-8$. The authors indicated these hyperparameters as the best ones.

        Their novel attention loss was implemented as a cross-entropy between the attention values and the rationale (the mean of attention losses for each attention head) by using an additional hyperparameter $\lambda$:
        \[\text{loss} = \text{cross-entropy} + \lambda \cdot \text{attention loss}.\]
        We explore the trade-off between their two losses (cross-entropy and attention loss) by varying $\lambda$ from 0.001 to 100 on a logarithmic scale, as suggested by the authors.
        Because our method considers the rationale binary (a token is either a rationale token or not), we also incorporated the rationales in BERT-HateXplain as binary, differently from their implementation, which uses the mean of the binary rationales (one for each annotator) as the rationale.
        Doing this was necessary for a fair comparison between the two methods.

        Even though we implement BERT-HateXplain with a few reasonable, justified modifications, our experimental results of their model are comparable to their paper's \citep{mathew_hatexplain_2021}, as pointed in \autoref{sec:comparison_with_hatexplain}.

\section{Additional Results}
    \label{appendix:additional_results}

    \subsection{Main Results}

        \begin{figure}[ht]
            \includegraphics[width=\linewidth]{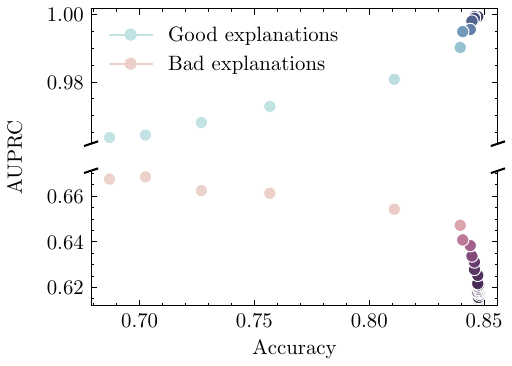}
            \caption{Trade-off between performance and plausibility on test data for originally good ($\text{AUPRC} = 1$) and originally bad ($\text{AUPRC} < 1$) explanations differently. The color scale is the same as the previous figures.}
            \label{fig:sample_level_auprc}
        \end{figure}

        \begin{figure}[ht]
            \includegraphics[width=\linewidth]{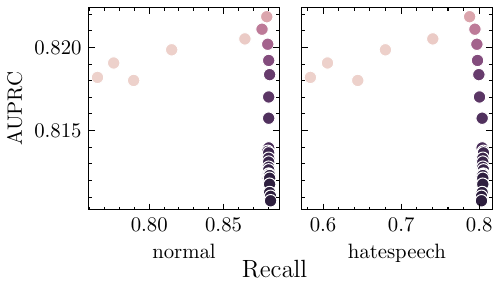}
            \caption{Trade-off between per class recall and plausibility on test data for DistilBERT and HateXplain dataset. The color scale is the same as the previous figures.}
            \label{fig:recall_auprc}
        \end{figure}

    \subsection{Results in Non-Binary Classification} \label{appendix:non_binary_results}

\begin{figure}[ht]
            \includegraphics[width=\linewidth]{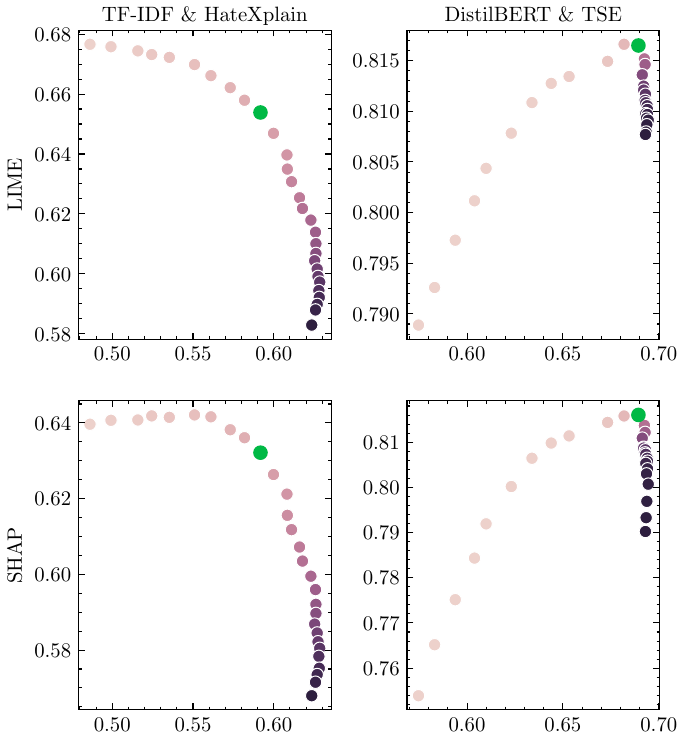}
            \caption{Trade-offs between performance (accuracy, x-axis) and plausibility (AUPRC, y-axis) for \texttt{hatexplain\_all} (i.e., with all labels, and with TF-IDF) and \texttt{tse\_all} (i.e., with all labels, and with DistilBERT) (test data). The number of random (negative) rationales is 2. The color scale is the same as the previous figures. We ignore the model with $w_1 = 0$ in all graphics as it is out of scale. Green dots are the models chosen to be analyzed more carefully.}
            \label{fig:three_classes_graphic}
        \end{figure}

        Sections \ref{sec:distilbert_and_hatexplain} and \ref{sec:other_experiments} present results for all datasets but are binary classification.
        As pointed out in \autoref{sec:datasets}, this procedure simplifies the learning task.
        Our methodology, however, is agnostic to the number of classes and can handle non-binary classification by default---we sum over any number of classes in \autoref{eq:contrastive_rationale_loss}.
        \autoref{fig:three_classes_graphic} presents the trade-off between accuracy and plausibility for \texttt{hatexplain\_all} (with TF-IDF) and \texttt{tse\_all} (with DistilBERT) (test data), i.e., with all the three labels, and a number of negative rationales of 2.
        The trade-off frontier shapes are similar to the binary classification, with similar conclusions from \autoref{sec:other_experiments}. However, different datasets lead to different absolute values.
        Finally, in a similar way to \autoref{sec:other_experiments}, \autoref{tab:three_classes_graphic} compares the original and chosen models, leading to similar conclusions: positive AUPRC improvement and a small decrease of performance.
        TSE had similar faithfulness results, while HateXplain had slightly worse faithfulness results.

        \begin{table*}[ht]
            \centering
            \caption{Comparison between the original model (cross-entropy only) and the chosen model (green dots on \autoref{fig:three_classes_graphic}) for each performance and explainability metric on test data. 
            ``rel.'' means relative variation. The column $w_1$ indicates the weight $w_1$ of the chosen model's cross-entropy loss during training. Number of negative rationales is 2.}
            \label{tab:three_classes_graphic}
            \begin{tabular}{lr|ccccc}
                \hline
                \textbf{Model}           & \textbf{$w_1$} & \textbf{Acc. \%} & \textbf{AUPRC \%} & \textbf{AUPRC rel. \%} & \textbf{Suff.} & \textbf{Comp.} \\
                \hline
                hatexplain\_all-lime-tf\_idf &  0.19 &  -3.17 &  7.09 &        12.16 &  -0.00 &  -0.06 \\
                hatexplain\_all-shap-tf\_idf &  0.19 &  -3.17 &  6.42 &        11.30 &  -0.00 &  -0.06 \\
                tse\_all-lime-distilbert    &  0.25 &  -0.37 &  0.88 &         1.09 &   0.01 &  -0.01 \\
                tse\_all-shap-distilbert    &  0.25 &  -0.37 &  2.58 &         3.26 &  -0.02 &  -0.00 \\
                \hline
            \end{tabular}
        \end{table*}

    \subsection{Results of Larger Models}
        \label{appendix:larger_models}
        
\begin{figure}[ht]
            \centering
            \includegraphics[width=\linewidth]{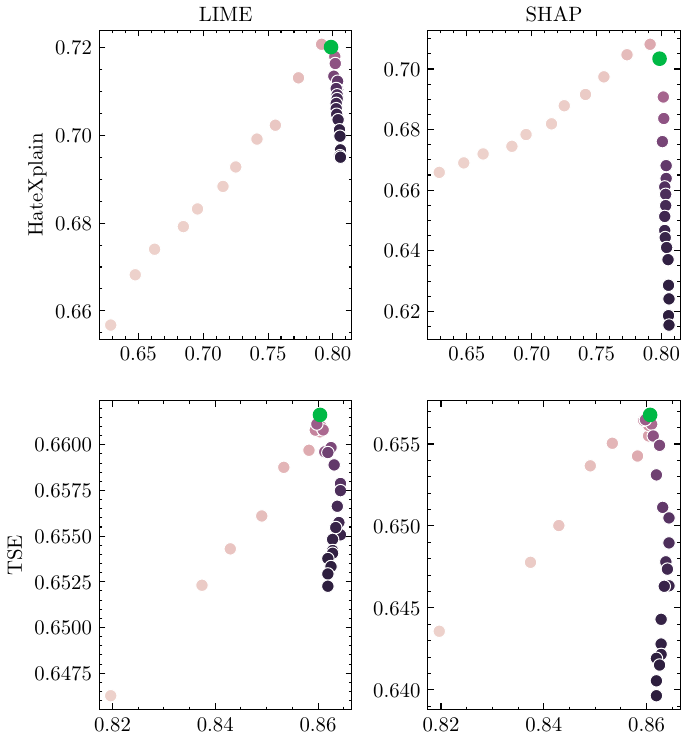}
            \caption{Trade-offs between performance (accuracy, x-axis) and plausibility (AUPRC, y-axis, in percentage (\%)) for BERT-Large with HateXplain and TSE (test data). The number of random (negative) rationales is 2, and the explainers are LIME and SHAP. The color scale is the same as the previous figures. We ignore the model with $w_1 = 0$ in all graphics as it is out of scale. Green dots are the models chosen to be analyzed more carefully.}
            \label{fig:bert_large_graphic}
        \end{figure}

        \autoref{sec:experiments} presents experiments with DistilBERT and BERT-Mini, which are small language model encoders. To further evaluate our methodology with a larger model, we performed a series of experiments with BERT-Large \cite{devlin_bert_2019}: datasets HateXplain and TSE, explainers LIME and SHAP, 2 negative rationales, BERT-Large without MLM fine-tuning.
        The shapes of the model frontiers (\autoref{fig:bert_large_graphic}) were similar to other language model frontiers of \autoref{fig:complete_graphic} in the main paper.
        Additionally, \autoref{tab:bert_large_graphic} compares the original and chosen models (in green).
        It reinforces our previous results regarding plausibility gain and minor performance degradation while improving or keeping faithfulness.
        We also highlight the existence of an experiment with BERT-Base \cite{devlin_bert_2019} in the baseline comparison, a larger model than DistilBERT and BERT-Mini used in the main experiments.

        \begin{table*}[ht]
            \centering
            \caption{Comparison between the original model (cross-entropy only) and the chosen model (green dots on \autoref{fig:bert_large_graphic}) for each performance and explainability metric on test data. 
            ``rel.'' means relative variation. The column $w_1$ indicates the weight $w_1$ of the chosen model's cross-entropy loss during training. Number of negative rationales is 2.}
            \label{tab:bert_large_graphic}
            \begin{tabular}{lr|ccccc}
                \hline
                \textbf{Model}           & \textbf{$w_1$} & \textbf{Acc. \%} & \textbf{AUPRC \%} & \textbf{AUPRC rel. \%} & \textbf{Suff.} & \textbf{Comp.} \\
                \hline
                hatexplain-lime-bert\_large &  0.33 &  -0.73 &  2.51 &         3.61 &  0.13 &   0.03 \\
                hatexplain-shap-bert\_large &  0.33 &  -0.73 &  8.79 &        14.29 &  0.12 &   0.06 \\
                tse-lime-bert\_large        &  0.30 &  -0.15 &  0.94 &         1.44 &  0.06 &  -0.01 \\
                tse-shap-bert\_large        &  0.43 &  -0.12 &  1.71 &         2.68 &  0.05 &  -0.00 \\
                \hline
            \end{tabular}
        \end{table*}

    \subsection{Out-of-Distribution Results}
        \label{appendix:out_of_distribution}

\begin{figure}[t!]
            \centering
            \includegraphics[width=\linewidth]{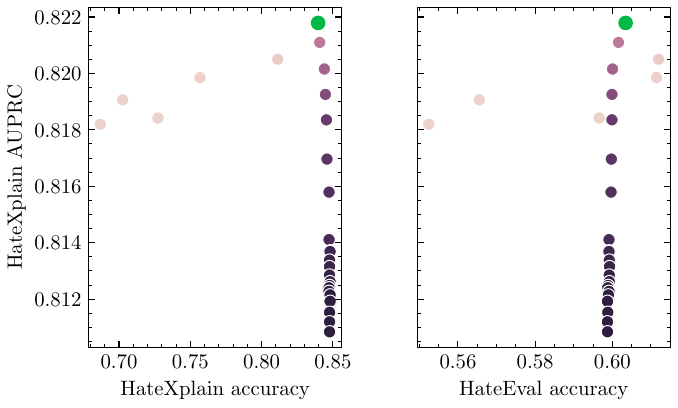}
            \caption{Trade-offs between (HateXplain and HatEval) performance and (HateXplain) plausibility with DistilBERT (test data). The number of random (negative) rationales is 2, and the explainer is LIME. The color scale is the same as the previous figures. We ignore the model with $w_1 = 0$ in all graphics as it is out of scale. Green dots are the model chosen to be analyzed more carefully.}
            \label{fig:out_of_distribution_graphic}
        \end{figure}

        To test out-of-distribution (OOD) performance, we additionally evaluated the DistilBERT trained on HateXplain (\autoref{sec:distilbert_and_hatexplain} of the main paper) on HatEval \cite{basile_semeval-2019_2019}, a similar dataset of hateful tweets but with a different data distribution (it focuses on hate speech against specific groups).
        We indeed observed an increase in OOD performance. The frontier shape of HatEval performance in \autoref{fig:out_of_distribution_graphic} is roughly similar to the frontier shape of HateXplain performance (in the same Figure and in \autoref{fig:accuracy_auprc}) but with the x-axis reversed (OOD performance increases with the plausibility, except for very small $w_1$ values). For the selected model (green dot in \autoref{fig:out_of_distribution_graphic}), while original accuracy decreases by 0.8\% and plausibility increases by approximately 1.1\%, the out-of-distribution performance also increases by 0.47\%. We also found it possible to increase by 0.97\% of plausibility and 1.32\% of OOD performance at the expense of a 3.64\% drop in original accuracy.

    \newpage

    \subsection{Other Results}
        \label{appendix:other_results}

        \begin{table*}[ht]
            \centering
            \caption{Examples of explanations of the hate speech class of the HateXplain dataset.
            Examples were selected based on the size and quality of the explanation and model predictions.
            The ``original'' explanation comes from the original model trained with cross-entropy loss only (\autoref{sec:distilbert_and_hatexplain}), while the ``selected'' explanation comes from the model with top-AUPRC studied in \autoref{sec:distilbert_and_hatexplain} (DistilBERT, HateXplain, LIME, 2 negative rationales).
            Green means a positive contribution to the model's prediction.
            The top tokens were selected for visualization purposes, and the number of tokens is the same as the original rationales.}
            \label{tab:more_bad_explanations}
            \begin{tabular}{c|l}
                \hline
                \textbf{Model} & \multicolumn{1}{c}{\textbf{Example}}                                                                                                                                                                                                                                                                                                                                                                                                                                                                                                                                                                               \\ \hline
                Original       & \colorbox{white}{bhad}\colorbox{movie}{bhabie}\colorbox{white}{we}\colorbox{movie}{hate}\colorbox{white}{your}\colorbox{white}{nasty}\colorbox{movie}{smelly}\colorbox{white}{*ss}                                                                                                                                                                                                                                                                                                                                                                                                                                 \\
                Selected       & \colorbox{white}{bhad}\colorbox{white}{bhabie}\colorbox{white}{we}\colorbox{movie}{hate}\colorbox{white}{your}\colorbox{movie}{nasty}\colorbox{movie}{smelly}\colorbox{white}{*ss}                                                                                                                                                                                                                                                                                                                                                                                                                                 \\ \hline
                Original       & \colorbox{white}{i}\colorbox{white}{tole}\colorbox{white}{sum}\colorbox{movie}{brillohaided}\colorbox{movie}{n****r}\colorbox{white}{bich}\colorbox{white}{off}\colorbox{white}{last}\colorbox{white}{nite}\colorbox{white}{cuz}\colorbox{white}{it}\colorbox{white}{try}\colorbox{white}{to}\colorbox{white}{get}\colorbox{white}{uppity}\colorbox{white}{wit}\colorbox{white}{me}                                                                                                                                                                                                                                \\
                Selected       & \colorbox{white}{i}\colorbox{white}{tole}\colorbox{white}{sum}\colorbox{white}{brillohaided}\colorbox{movie}{n****r}\colorbox{movie}{bich}\colorbox{white}{off}\colorbox{white}{last}\colorbox{white}{nite}\colorbox{white}{cuz}\colorbox{white}{it}\colorbox{white}{try}\colorbox{white}{to}\colorbox{white}{get}\colorbox{white}{uppity}\colorbox{white}{wit}\colorbox{white}{me}                                                                                                                                                                                                                                \\ \hline
                Original       & \colorbox{movie}{london}\colorbox{white}{mayor}\colorbox{white}{s*ckdiq}\colorbox{white}{kahn}\colorbox{white}{has}\colorbox{white}{room}\colorbox{white}{for}\colorbox{white}{a}\colorbox{white}{fellow}\colorbox{movie}{m**zie}\colorbox{white}{terrorist}                                                                                                                                                                                                                                                                                                                                                       \\
                Selected       & \colorbox{white}{london}\colorbox{white}{mayor}\colorbox{white}{s*ckdiq}\colorbox{white}{kahn}\colorbox{white}{has}\colorbox{white}{room}\colorbox{white}{for}\colorbox{white}{a}\colorbox{white}{fellow}\colorbox{movie}{m**zie}\colorbox{movie}{terrorist}                                                                                                                                                                                                                                                                                                                                                       \\ \hline
                Original       & \begin{tabular}[c]{@{}l@{}}\colorbox{white}{no}\colorbox{white}{<number>}\colorbox{white}{million}\colorbox{white}{white}\colorbox{white}{children}\colorbox{white}{r***d}\colorbox{white}{by}\colorbox{movie}{p*ki}\colorbox{white}{muslims}\colorbox{white}{<number>}\colorbox{white}{million}\colorbox{white}{p*kis}\\\colorbox{white}{in}\colorbox{white}{the}\colorbox{movie}{uk}\colorbox{white}{for} \colorbox{white}{their}\colorbox{white}{crime}\colorbox{white}{the}\colorbox{white}{whole}\colorbox{white}{community}\colorbox{white}{must}\colorbox{white}{be}\colorbox{white}{expelled}\end{tabular} \\
                Selected       & \begin{tabular}[c]{@{}l@{}}\colorbox{white}{no}\colorbox{white}{<number>}\colorbox{white}{million}\colorbox{white}{white}\colorbox{white}{children}\colorbox{white}{r***d}\colorbox{white}{by}\colorbox{movie}{p*ki}\colorbox{white}{muslims}\colorbox{white}{<number>}\colorbox{white}{million}\colorbox{movie}{p*kis}\\\colorbox{white}{in}\colorbox{white}{the}\colorbox{white}{uk}\colorbox{white}{for} \colorbox{white}{their}\colorbox{white}{crime}\colorbox{white}{the}\colorbox{white}{whole}\colorbox{white}{community}\colorbox{white}{must}\colorbox{white}{be}\colorbox{white}{expelled}\end{tabular} \\ \hline
                Original       & \begin{tabular}[c]{@{}l@{}}\colorbox{white}{the}\colorbox{white}{russians}\colorbox{white}{buying}\colorbox{white}{that}\colorbox{white}{and}\colorbox{white}{purging}\colorbox{white}{all}\colorbox{white}{the}\colorbox{white}{blue}\colorbox{white}{haired}\colorbox{white}{d*kes}\colorbox{white}{was}\colorbox{white}{glorious}\\ \colorbox{white}{that}\colorbox{white}{may}\colorbox{white}{well}\colorbox{white}{be}\colorbox{white}{the}\colorbox{white}{genesis}\colorbox{white}{of}\colorbox{white}{the}\colorbox{movie}{russian}\colorbox{white}{bot}\colorbox{white}{narrative}\end{tabular}          \\
                Selected       & \begin{tabular}[c]{@{}l@{}}\colorbox{white}{the}\colorbox{white}{russians}\colorbox{white}{buying}\colorbox{white}{that}\colorbox{white}{and}\colorbox{white}{purging}\colorbox{white}{all}\colorbox{white}{the}\colorbox{white}{blue}\colorbox{white}{haired}\colorbox{movie}{d*kes}\colorbox{white}{was}\colorbox{white}{glorious}\\ \colorbox{white}{that}\colorbox{white}{may}\colorbox{white}{well}\colorbox{white}{be}\colorbox{white}{the}\colorbox{white}{genesis}\colorbox{white}{of}\colorbox{white}{the}\colorbox{white}{russian}\colorbox{white}{bot}\colorbox{white}{narrative}\end{tabular}          \\ \hline
                Original       & \begin{tabular}[c]{@{}l@{}}\colorbox{white}{give}\colorbox{white}{a}\colorbox{movie}{n****r}\colorbox{white}{an}\colorbox{white}{inch}\colorbox{white}{they}\colorbox{white}{take}\colorbox{white}{a}\colorbox{white}{mile}\colorbox{white}{lol}\colorbox{movie}{r*tarded}\colorbox{white}{creatures}\colorbox{white}{they}\colorbox{movie}{f*ck}\\ \colorbox{white}{themselves}\colorbox{white}{over}\colorbox{white}{every}\colorbox{white}{time}\end{tabular}                                                                                                                                                   \\
                Selected       & \begin{tabular}[c]{@{}l@{}}\colorbox{white}{give}\colorbox{white}{a}\colorbox{movie}{n****r}\colorbox{white}{an}\colorbox{white}{inch}\colorbox{white}{they}\colorbox{white}{take}\colorbox{white}{a}\colorbox{white}{mile}\colorbox{white}{lol}\colorbox{movie}{r*tarded}\colorbox{movie}{creatures}\colorbox{white}{they}\colorbox{white}{f*ck}\\ \colorbox{white}{themselves}\colorbox{white}{over}\colorbox{white}{every}\colorbox{white}{time}\end{tabular}                                                                                                                                                   \\ \hline
                Original       & \begin{tabular}[c]{@{}l@{}}\colorbox{white}{and}\colorbox{white}{if}\colorbox{white}{u}\colorbox{white}{have}\colorbox{white}{to}\colorbox{white}{drink}\colorbox{white}{and}\colorbox{white}{drive}\colorbox{white}{make}\colorbox{white}{sure}\colorbox{white}{u}\colorbox{white}{drive}\colorbox{white}{home}\colorbox{white}{through}\colorbox{white}{as}\colorbox{movie}{many}\\ \colorbox{movie}{n****r}\colorbox{white}{gh*ttos}\colorbox{white}{as}\colorbox{white}{possible}\colorbox{white}{[emoji]}\end{tabular}                                                                                        \\
                Selected       & \begin{tabular}[c]{@{}l@{}}\colorbox{white}{and}\colorbox{white}{if}\colorbox{white}{u}\colorbox{white}{have}\colorbox{white}{to}\colorbox{white}{drink}\colorbox{white}{and}\colorbox{white}{drive}\colorbox{white}{make}\colorbox{white}{sure}\colorbox{white}{u}\colorbox{white}{drive}\colorbox{white}{home}\colorbox{white}{through}\colorbox{white}{as}\colorbox{white}{many}\\ \colorbox{movie}{n****r}\colorbox{movie}{gh*ttos}\colorbox{white}{as}\colorbox{white}{possible}\colorbox{white}{[emoji]}\end{tabular}                                                                                        \\ \hline
            \end{tabular}
        \end{table*}

        \begin{table*}[ht]
            \centering
            \caption{Examples of explanations of the Tweet Sentiment Extraction dataset.
            Examples were selected based on the size and quality of the explanation and model predictions.
            The ``original'' explanation (LIME) comes from the original DistilBERT model trained with cross-entropy loss only (\autoref{sec:other_experiments}), while the ``selected'' explanation comes from the selected model with a green dot (\autoref{sec:other_experiments}, \autoref{fig:complete_graphic}) (2 negative rationales).
            Green means a positive contribution to the model's prediction.
            The top tokens were selected for visualization purposes, and the number of tokens is the same as the original rationales.}
            \label{tab:more_more_bad_explanations}
            \begin{tabular}{c|c|l}
                \hline
                \textbf{Label}            & \textbf{Model} & \multicolumn{1}{c}{\textbf{Example}}                                                                                                                                                                                                                                                  \\ \hline
                \multirow{2}{*}{positive} & Original       & \colorbox{white}{in}\colorbox{white}{rye}\colorbox{white}{.}\colorbox{white}{.}\colorbox{white}{happy}\colorbox{white}{mothers}\colorbox{white}{day}\colorbox{white}{mums}\colorbox{movie}{ily}\colorbox{white}{mummy}\colorbox{white}{lol}                                           \\
                                        & Selected       & \colorbox{white}{in}\colorbox{white}{rye}\colorbox{white}{.}\colorbox{white}{.}\colorbox{movie}{happy}\colorbox{white}{mothers}\colorbox{white}{day}\colorbox{white}{mums}\colorbox{white}{ily}\colorbox{white}{mummy}\colorbox{white}{lol}                                           \\ \hline
                \multirow{2}{*}{positive} & Original       & \colorbox{white}{I}\colorbox{movie}{`}\colorbox{white}{ll}\colorbox{white}{try}\colorbox{white}{that}\colorbox{white}{,}\colorbox{white}{thanks}                                                                                                                                      \\
                                        & Selected       & \colorbox{white}{I}\colorbox{white}{`}\colorbox{white}{ll}\colorbox{white}{try}\colorbox{white}{that}\colorbox{white}{,}\colorbox{movie}{thanks}                                                                                                                                      \\ \hline
                \multirow{2}{*}{positive} & Original       & \colorbox{white}{LOVE}\colorbox{white}{your}\colorbox{movie}{show}\colorbox{white}{!}                                                                                                                                                                                                 \\
                                        & Selected       & \colorbox{movie}{LOVE}\colorbox{white}{your}\colorbox{white}{show}\colorbox{white}{!}                                                                                                                                                                                                 \\ \hline
                \multirow{2}{*}{positive} & Original       & \colorbox{white}{\_}\colorbox{white}{O}\colorbox{white}{\_}\colorbox{white}{ASH}\colorbox{white}{I}\colorbox{white}{do}\colorbox{white}{too}\colorbox{white}{plus}\colorbox{white}{more}\colorbox{white}{happy}\colorbox{white}{mothers}\colorbox{white}{day}\colorbox{movie}{Sweety} \\
                                        & Selected       & \colorbox{white}{\_}\colorbox{white}{O}\colorbox{white}{\_}\colorbox{white}{ASH}\colorbox{white}{I}\colorbox{white}{do}\colorbox{white}{too}\colorbox{white}{plus}\colorbox{white}{more}\colorbox{movie}{happy}\colorbox{white}{mothers}\colorbox{white}{day}\colorbox{white}{Sweety} \\ \hline
                \multirow{2}{*}{positive} & Original       & \colorbox{white}{hopefully}\colorbox{white}{today}\colorbox{white}{will}\colorbox{white}{work}\colorbox{white}{in}\colorbox{white}{our}\colorbox{movie}{favor}                                                                                                                        \\
                                        & Selected       & \colorbox{movie}{hopefully}\colorbox{white}{today}\colorbox{white}{will}\colorbox{white}{work}\colorbox{white}{in}\colorbox{white}{our}\colorbox{white}{favor}                                                                                                                        \\ \hline
                \multirow{2}{*}{positive} & Original       & \colorbox{movie}{Rachmaninoff}\colorbox{white}{makes}\colorbox{white}{me}\colorbox{white}{a}\colorbox{white}{happy}\colorbox{white}{panda}\colorbox{white}{.}                                                                                                                         \\
                                        & Selected       & \colorbox{white}{Rachmaninoff}\colorbox{white}{makes}\colorbox{white}{me}\colorbox{white}{a}\colorbox{movie}{happy}\colorbox{white}{panda}\colorbox{white}{.}                                                                                                                         \\ \hline
                \multirow{2}{*}{positive} & Original       & \colorbox{white}{You}\colorbox{white}{must}\colorbox{white}{like}\colorbox{white}{my}\colorbox{movie}{song}\colorbox{white}{.}                                                                                                                                                        \\
                                        & Selected       & \colorbox{white}{You}\colorbox{white}{must}\colorbox{movie}{like}\colorbox{white}{my}\colorbox{white}{song}\colorbox{white}{.}                                                                                                                                                        \\ \hline
                \multirow{2}{*}{negative} & Original       & \colorbox{white}{\_}\colorbox{movie}{[user]}\colorbox{white}{aww}\colorbox{white}{that}\colorbox{white}{sucks}                                                                                                                                                                        \\
                                        & Selected       & \colorbox{white}{\_}\colorbox{white}{[user]}\colorbox{white}{aww}\colorbox{white}{that}\colorbox{movie}{sucks}                                                                                                                                                                        \\ \hline
                \multirow{2}{*}{positive} & Original       & \colorbox{white}{Digging}\colorbox{white}{a}\colorbox{movie}{downloaded}\colorbox{white}{film}\colorbox{white}{with}\colorbox{white}{mi}\colorbox{white}{familia}\colorbox{white}{.}\colorbox{white}{We}\colorbox{white}{love}\colorbox{white}{iTunes}                                \\
                                        & Selected       & \colorbox{white}{Digging}\colorbox{white}{a}\colorbox{white}{downloaded}\colorbox{white}{film}\colorbox{white}{with}\colorbox{white}{mi}\colorbox{white}{familia}\colorbox{white}{.}\colorbox{white}{We}\colorbox{movie}{love}\colorbox{white}{iTunes}                                \\ \hline
                \multirow{2}{*}{positive} & Original       & \colorbox{white}{Happy}\colorbox{white}{Mommy}\colorbox{movie}{Day}                                                                                                                                                                                                                   \\
                                        & Selected       & \colorbox{movie}{Happy}\colorbox{white}{Mommy}\colorbox{white}{Day}                                                                                                                                                                                                                   \\ \hline
            \end{tabular}
        \end{table*}

        \begin{figure}[ht]
            \includegraphics[width=\linewidth]{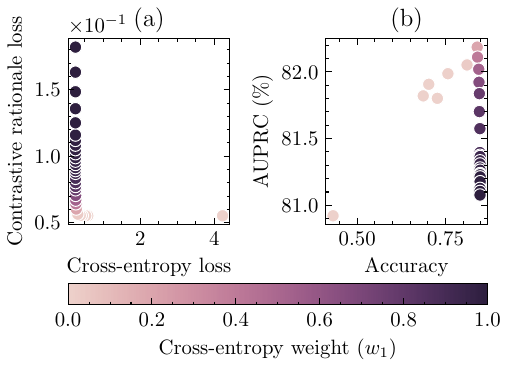}
            \caption{(a) Trade-off between the two losses on the training data. (b) Trade-off between accuracy and plausibility on the test data. The color scale represents the cross-entropy weight $w_1$ (\autoref{sec:trade_off}). We include the model with $w_1 = 0$.}
            \label{fig:accuracy_auprc_all}
        \end{figure}

        \begin{figure}[ht]
            \centering
            \includegraphics[width=\linewidth]{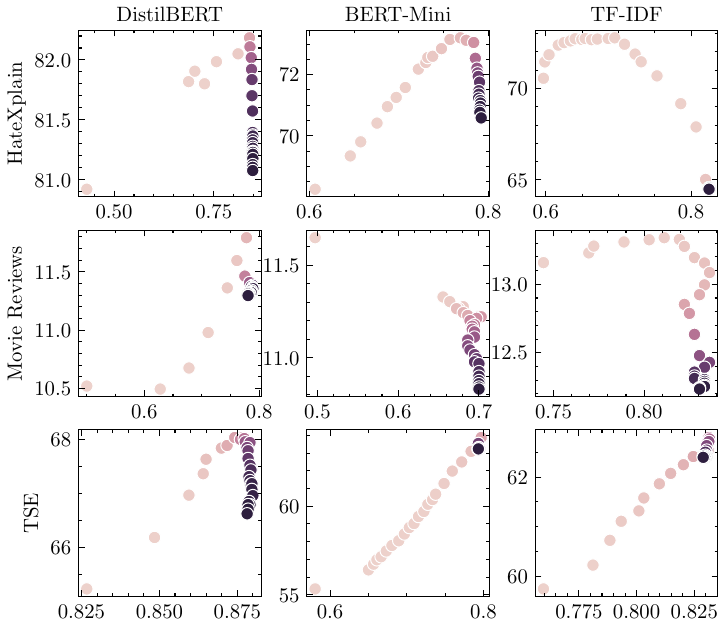}
            \caption{Trade-offs between performance (accuracy, x-axis) and plausibility (AUPRC, y-axis, in percentage (\%)) for all models and datasets (test data). The number of random (negative) rationales is 2, and the explainer is LIME. The color scale is the same as the previous figures. We include the model with $w_1 = 0$ in all graphics.}
            \label{fig:complete_graphic_all}
        \end{figure}

        \begin{figure}[ht]
            \centering
            \includegraphics[width=\linewidth]{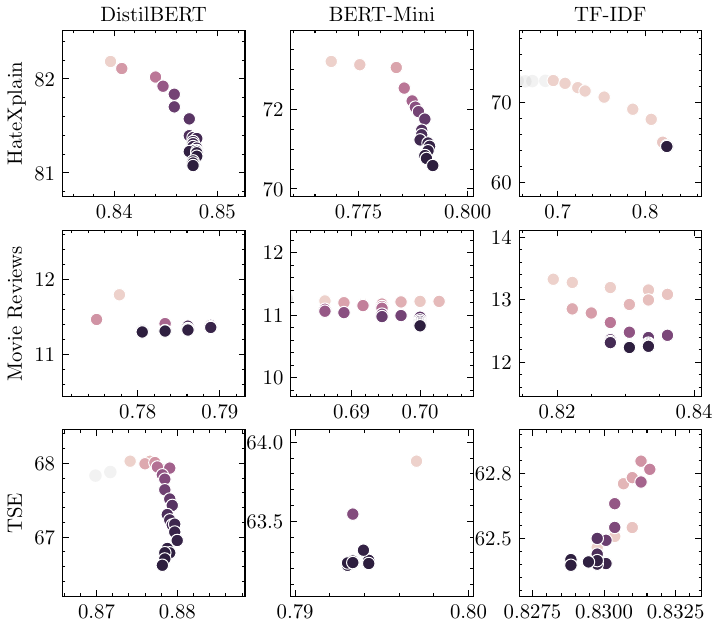}
            \caption{Trade-offs between performance (accuracy, x-axis) and plausibility (AUPRC, y-axis, in percentage (\%)) for all models and datasets (test data). The number of random (negative) rationales is 2, and the explainer is LIME. The color scale is the same as the previous figures. The data scale is equal between x- and y-axes, and a few out-of-scale points are gray or removed.}
            \label{fig:complete_graphic_desired}
        \end{figure}

        \begin{figure*}[ht]
            \centering
            \includegraphics[width=0.8\linewidth]{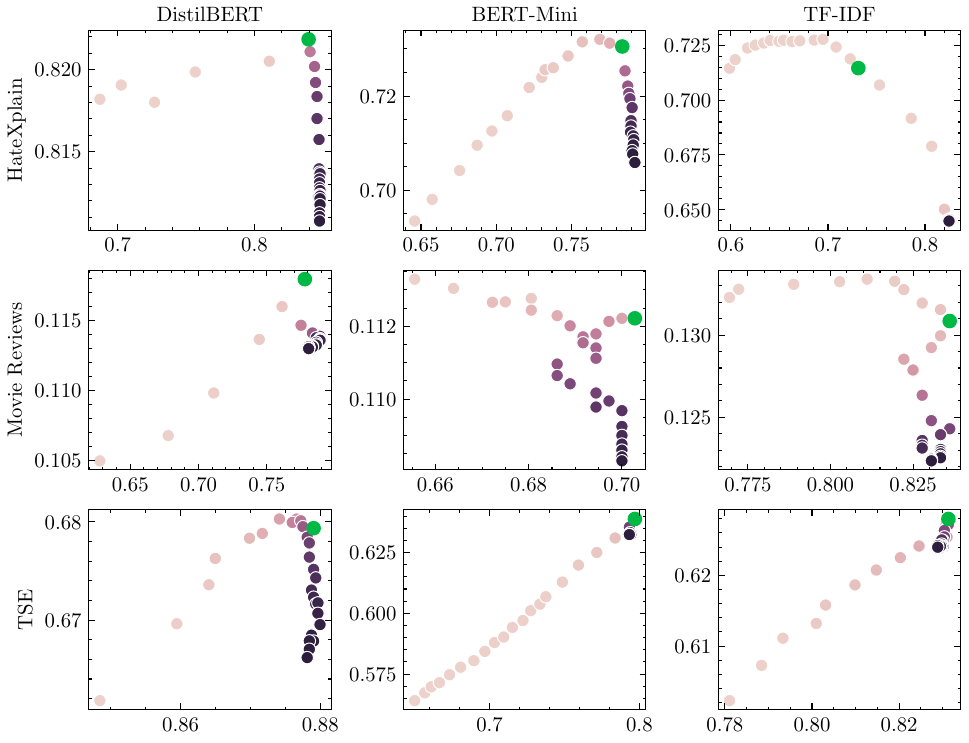}
            \caption{Trade-offs between performance (accuracy, x-axis) and plausibility (AUPRC, y-axis) for all models and datasets (test data). The number of random (negative) rationales is 2, and the explainer is LIME. The color scale is the same as the previous figures. We ignore the model with $w_1 = 0$ in all graphics as it is out of scale. Green dots are the models chosen to be analyzed more carefully.}
            \label{fig:complete_graphic_lime_2}
        \end{figure*}

        \begin{figure*}[ht]
            \centering
            \includegraphics[width=0.8\linewidth]{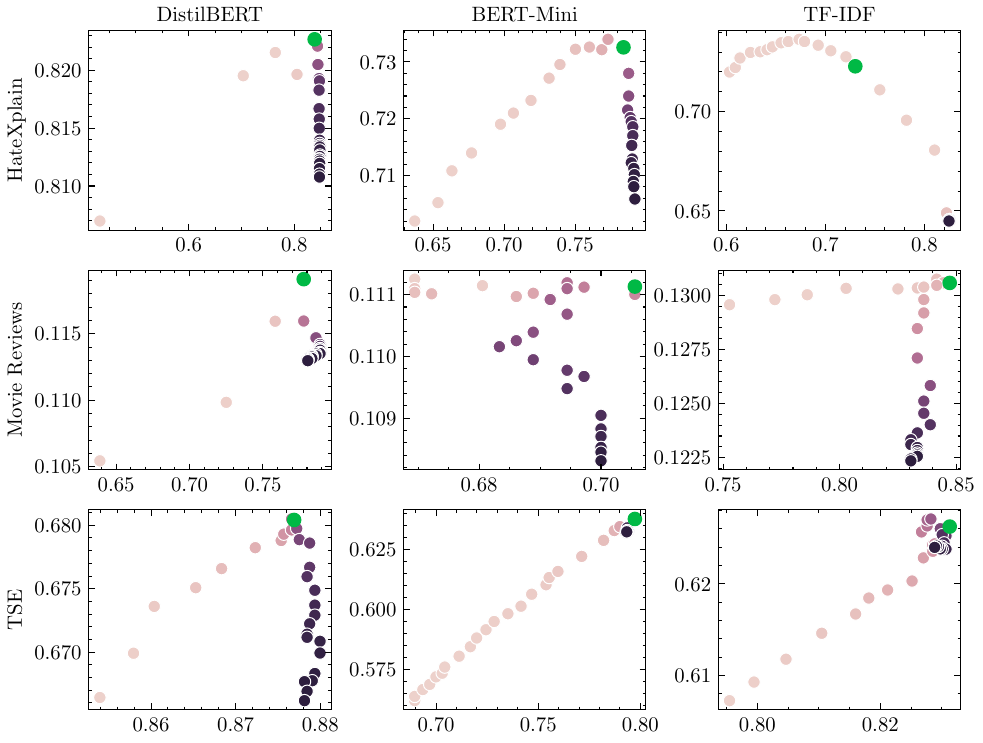}
            \caption{Trade-offs between performance (accuracy, x-axis) and plausibility (AUPRC, y-axis) for all models and datasets (test data). The number of random (negative) rationales is 5, and the explainer is LIME. The color scale is the same as the previous figures. We ignore the model with $w_1 = 0$ in all graphics as it is out of scale. Green dots are the models chosen to be analyzed more carefully.}
            \label{fig:complete_graphic_lime_5}
        \end{figure*}

        \begin{figure*}[ht]
            \centering
            \includegraphics[width=0.8\linewidth]{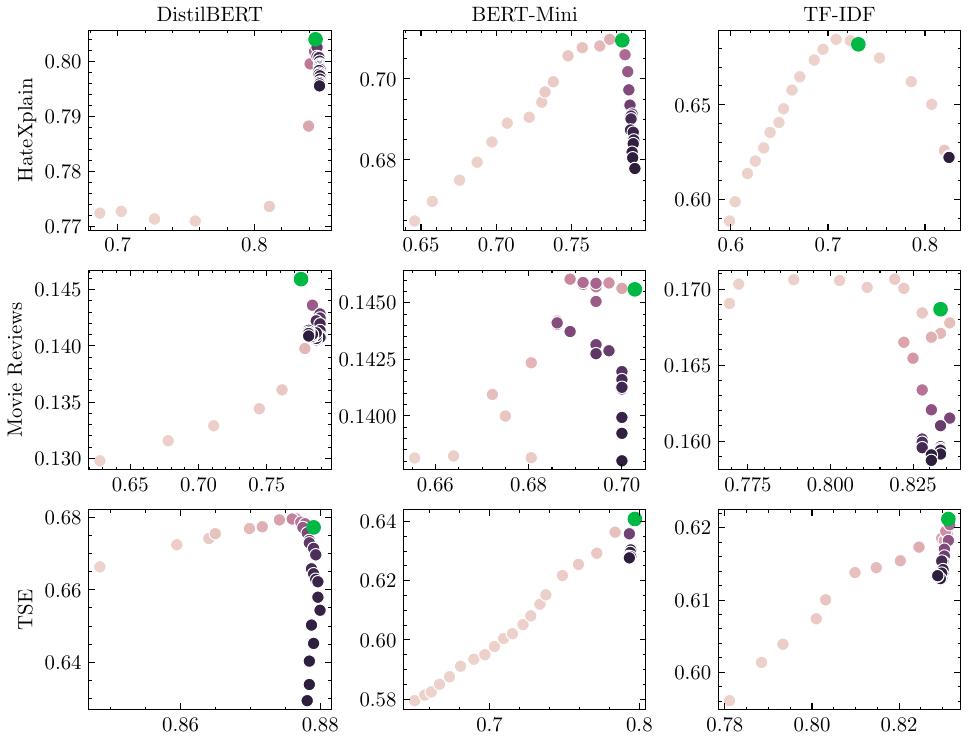}
            \caption{Trade-offs between performance (accuracy, x-axis) and plausibility (AUPRC, y-axis) for all models and datasets (test data). The number of random (negative) rationales is 2, and the explainer is SHAP. The color scale is the same as the previous figures. We ignore the model with $w_1 = 0$ in all graphics as it is out of scale. Green dots are the models chosen to be analyzed more carefully.}
            \label{fig:complete_graphic_shap_2}
        \end{figure*}

        \begin{figure*}[ht]
            \centering
            \includegraphics[width=0.8\linewidth]{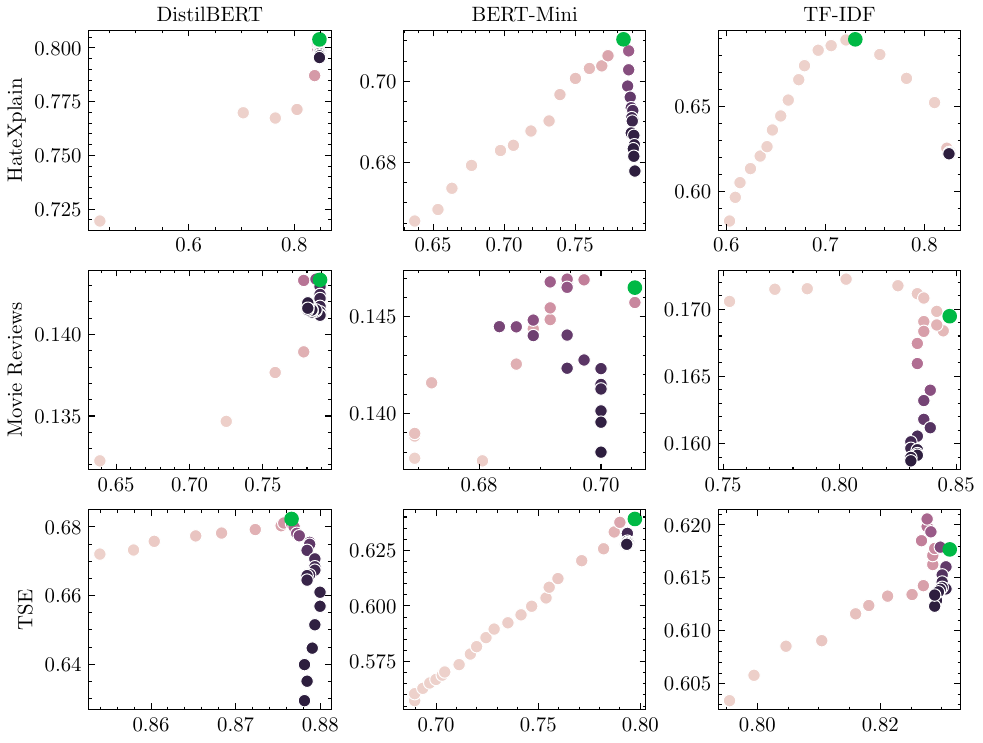}
            \caption{Trade-offs between performance (accuracy, x-axis) and plausibility (AUPRC, y-axis) for all models and datasets (test data). The number of random (negative) rationales is 5, and the explainer is SHAP. The color scale is the same as the previous figures. We ignore the model with $w_1 = 0$ in all graphics as it is out of scale. Green dots are the models chosen to be analyzed more carefully.}
            \label{fig:complete_graphic_shap_5}
        \end{figure*}

        \begin{figure*}[ht]
            \centering
            \includegraphics[width=0.8\linewidth]{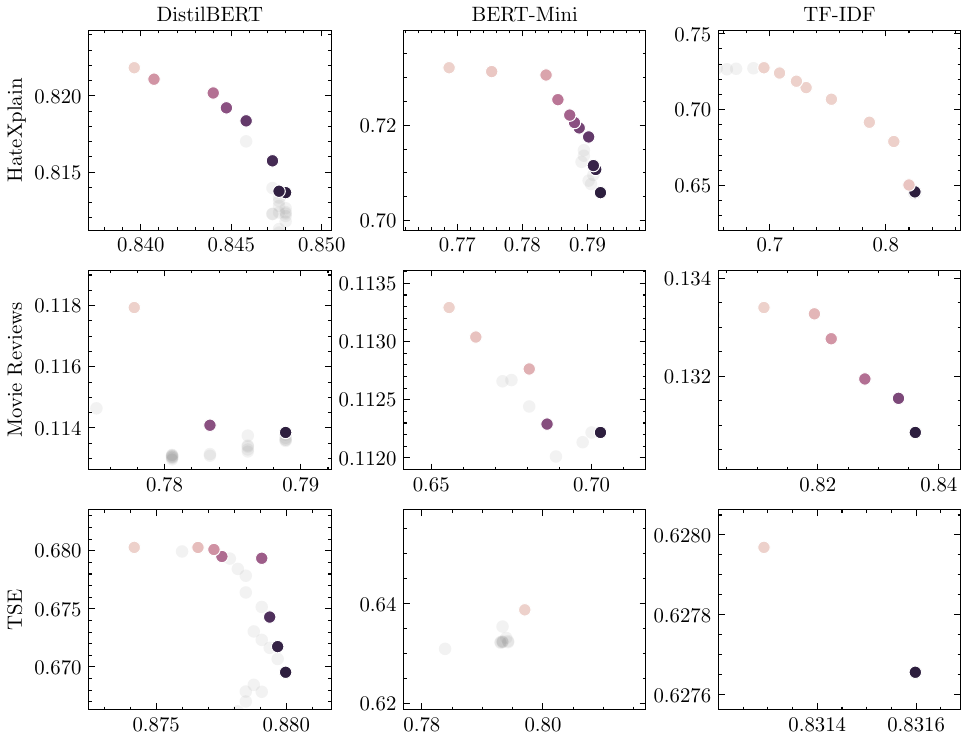}
            \caption{Pareto-frontier of trade-offs between performance (accuracy, x-axis) and plausibility (AUPRC, y-axis) for all models and datasets (test data). The number of random (negative) rationales is 2, and the explainer is LIME. The color scale is the same as the previous figures. Gray dots are models not on the Pareto-frontier. We ignore the model with $w_1 = 0$ in all graphics as it is out of scale.}
            \label{fig:complete_graphic_lime_2_frontier}
        \end{figure*}

        \begin{figure*}[ht]
            \centering
            \includegraphics[width=0.8\linewidth]{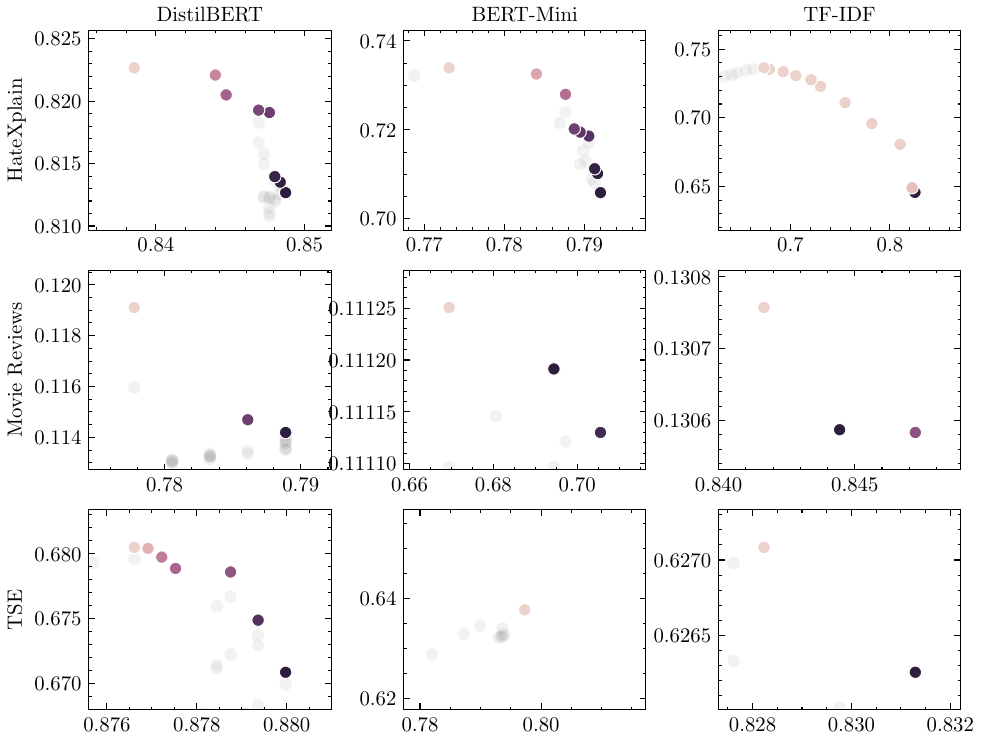}
            \caption{Pareto-frontier of trade-offs between performance (accuracy, x-axis) and plausibility (AUPRC, y-axis) for all models and datasets (test data). The number of random (negative) rationales is 5, and the explainer is LIME. The color scale is the same as the previous figures. Gray dots are models not on the Pareto-frontier. We ignore the model with $w_1 = 0$ in all graphics as it is out of scale.}
            \label{fig:complete_graphic_lime_5_frontier}
        \end{figure*}

        \begin{figure*}[ht]
            \centering
            \includegraphics[width=0.8\linewidth]{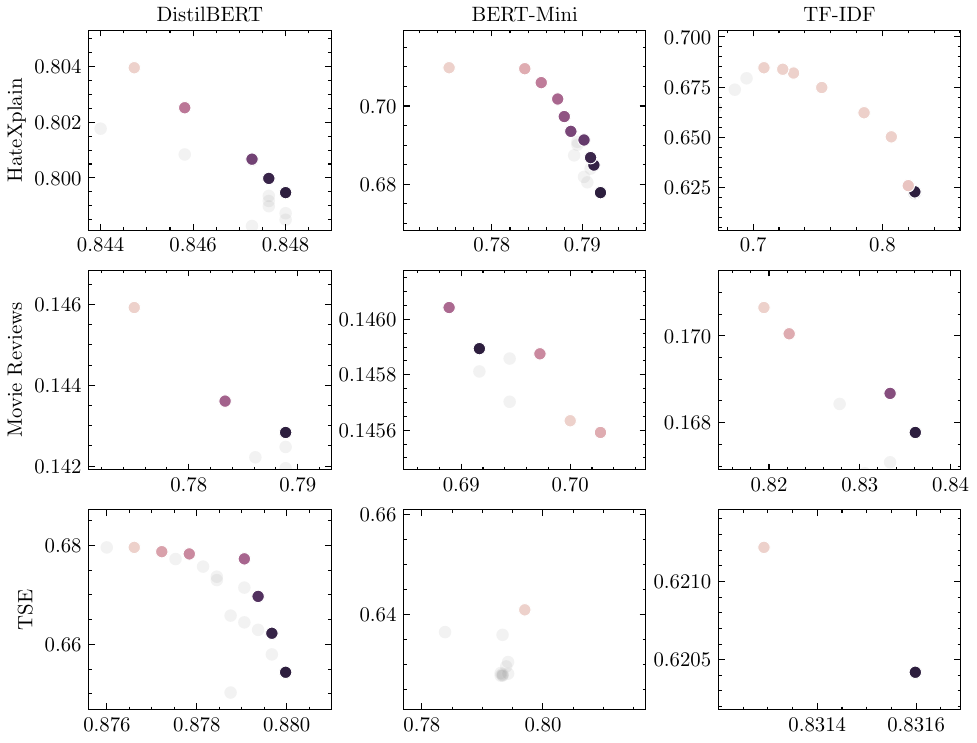}
            \caption{Pareto-frontier of trade-offs between performance (accuracy, x-axis) and plausibility (AUPRC, y-axis) for all models and datasets (test data). The number of random (negative) rationales is 2, and the explainer is SHAP. The color scale is the same as the previous figures. Gray dots are models not on the Pareto-frontier. We ignore the model with $w_1 = 0$ in all graphics as it is out of scale.}
            \label{fig:complete_graphic_shap_2_frontier}
        \end{figure*}

        \begin{figure*}[ht]
            \centering
            \includegraphics[width=0.8\linewidth]{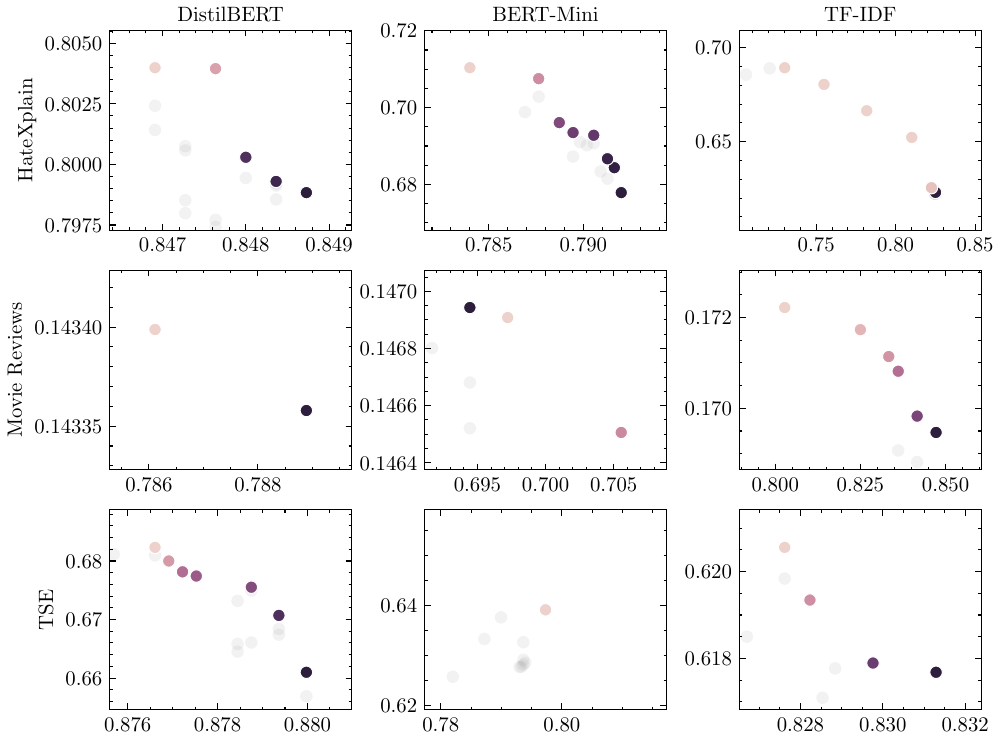}
            \caption{Pareto-frontier of trade-offs between performance (accuracy, x-axis) and plausibility (AUPRC, y-axis) for all models and datasets (test data). The number of random (negative) rationales is 5, and the explainer is SHAP. The color scale is the same as the previous figures. Gray dots are models not on the Pareto-frontier. We ignore the model with $w_1 = 0$ in all graphics as it is out of scale.}
            \label{fig:complete_graphic_shap_5_frontier}
        \end{figure*}

        \begin{table*}[ht]
            \centering
            \caption{Comparison between the original model (cross-entropy only) and the chosen model (green dots on Figures \ref{fig:complete_graphic}, \ref{fig:complete_graphic_lime_5}, \ref{fig:complete_graphic_shap_2}, \ref{fig:complete_graphic_shap_5}) for each performance and explainability metric on test data. 
            ``rel.'' means relative variation. The column $w_1$ indicates the weight $w_1$ of the chosen model's cross-entropy loss during training.}
            \label{tab:complete_graphic_all}
            \begin{tabular}{lr|ccccc}
                \hline
                \textbf{Model}           & \textbf{$w_1$} & \textbf{Acc. \%} & \textbf{AUPRC \%} & \textbf{AUPRC rel. \%} & \textbf{Suff.} & \textbf{Comp.} \\
                \hline
                hatexplain-lime-distilbert-2    &   0.20 &  -0.80 &  1.11 &         1.37 &   0.25 &  -0.03 \\
                hatexplain-shap-distilbert-2    &   0.67 &  -0.29 &  0.85 &         1.06 &   0.15 &  -0.01 \\
                hatexplain-lime-distilbert-5    &   0.25 &  -0.91 &  1.19 &         1.47 &   0.25 &  -0.03 \\
                hatexplain-shap-distilbert-5    &   0.80 &   0.00 &  0.85 &         1.06 &   0.14 &  -0.01 \\
                hatexplain-lime-bert\_mini-2     &   0.29 &  -0.84 &  2.46 &         3.49 &   0.40 &  -0.05 \\
                hatexplain-shap-bert\_mini-2     &   0.29 &  -0.84 &  3.17 &         4.67 &   0.40 &  -0.05 \\
                hatexplain-lime-bert\_mini-5     &   0.37 &  -0.80 &  2.67 &         3.78 &   0.41 &  -0.04 \\
                hatexplain-shap-bert\_mini-5     &   0.37 &  -0.80 &  3.25 &         4.80 &   0.40 &  -0.05 \\
                hatexplain-lime-tf\_idf-2        &  0.002 &  -9.35 &  6.96 &        10.79 &   0.13 &  -0.10 \\
                hatexplain-shap-tf\_idf-2        &  0.002 &  -9.35 &  5.98 &         9.60 &   0.13 &  -0.09 \\
                hatexplain-lime-tf\_idf-5        &  0.002 &  -9.45 &  7.79 &        12.08 &   0.13 &  -0.10 \\
                hatexplain-shap-tf\_idf-5        &  0.002 &  -9.45 &  6.71 &        10.79 &   0.14 &  -0.10 \\
                movie\_reviews-lime-distilbert-2 &   0.12 &  -0.28 &  0.50 &         4.39 &   0.25 &  -0.05 \\
                movie\_reviews-shap-distilbert-2 &   0.36 &  -0.56 &  0.50 &         3.58 &   0.13 &  -0.02 \\
                movie\_reviews-lime-distilbert-5 &   0.15 &  -0.28 &  0.61 &         5.43 &   0.25 &  -0.02 \\
                movie\_reviews-shap-distilbert-5 &   0.81 &   0.83 &  0.17 &         1.23 &   0.04 &   0.00 \\
                movie\_reviews-lime-bert\_mini-2  &   0.26 &   0.28 &  0.39 &         3.61 &   0.00 &  -0.02 \\
                movie\_reviews-shap-bert\_mini-2  &   0.26 &   0.28 &  0.76 &         5.49 &  -0.01 &  -0.02 \\
                movie\_reviews-lime-bert\_mini-5  &   0.43 &   0.56 &  0.28 &         2.60 &   0.02 &  -0.01 \\
                movie\_reviews-shap-bert\_mini-5  &   0.43 &   0.56 &  0.85 &         6.16 &   0.01 &  -0.01 \\
                movie\_reviews-lime-tf\_idf-2     &   0.09 &   0.56 &  0.85 &         6.95 &  -0.00 &   0.01 \\
                movie\_reviews-shap-tf\_idf-2     &   0.07 &   0.28 &  0.99 &         6.26 &   0.01 &   0.01 \\
                movie\_reviews-lime-tf\_idf-5     &   0.10 &   1.67 &  0.82 &         6.73 &  -0.02 &   0.01 \\
                movie\_reviews-shap-tf\_idf-5     &   0.10 &   1.67 &  1.07 &         6.77 &  -0.02 &   0.02 \\
                tse-lime-distilbert-2           &   0.64 &   0.09 &  1.32 &         1.98 &   0.05 &  -0.00 \\
                tse-shap-distilbert-2           &   0.64 &   0.09 &  4.79 &         7.61 &   0.00 &   0.02 \\
                tse-lime-distilbert-5           &   0.51 &  -0.12 &  1.42 &         2.14 &   0.07 &   0.00 \\
                tse-shap-distilbert-5           &   0.36 &  -0.15 &  5.29 &         8.41 &   0.04 &   0.03 \\
                tse-lime-bert\_mini-2            &   0.19 &   0.37 &  0.64 &         1.01 &   0.06 &   0.01 \\
                tse-shap-bert\_mini-2            &   0.19 &   0.37 &  1.31 &         2.09 &   0.06 &   0.01 \\
                tse-lime-bert\_mini-5            &   0.43 &   0.40 &  0.54 &         0.85 &   0.06 &   0.01 \\
                tse-shap-bert\_mini-5            &   0.43 &   0.40 &  1.14 &         1.81 &   0.05 &   0.01 \\
                tse-lime-tf\_idf-2               &   0.42 &   0.24 &  0.40 &         0.64 &   0.01 &  -0.02 \\
                tse-shap-tf\_idf-2               &   0.42 &   0.24 &  0.78 &         1.28 &   0.01 &  -0.02 \\
                tse-lime-tf\_idf-5               &   0.75 &   0.24 &  0.23 &         0.36 &   0.00 &  -0.01 \\
                tse-shap-tf\_idf-5               &   0.75 &   0.24 &  0.43 &         0.70 &   0.00 &  -0.01 \\
                \hline
            \end{tabular}
        \end{table*}